\RequirePackage[svgnames,table]{xcolor}
\PassOptionsToPackage{hyphens}{url}
\documentclass[11pt,letterpaper]{yalearxiv}

\usepackage{latexsym}
\usepackage[authoryear,round]{natbib}
\renewcommand\cite{\citep}

\usepackage{longtable}
\usepackage{multirow}
\usepackage{algorithm}
\usepackage{algorithmic}
\usepackage{subcaption}
\usepackage{fancyvrb}
\usepackage{newfloat}
\usepackage{listings}

\ifdefined\XeTeXversion
  \usepackage{newunicodechar}
  \newunicodechar{“}{\textquotedblleft}
  \newunicodechar{”}{\textquotedblright}
  \newunicodechar{–}{\textendash}
  \newcount\SPDocReaderCompressionCompatibility
  
  \let\pdfoptionpdfminorversion\SPDocReaderCompressionCompatibility
  
  \long\def\pdfglyphtounicode#1#2{}
\else
  \ifdefined\pdfoptionpdfminorversion\else
    \ifdefined\pdfminorversion
      \let\pdfoptionpdfminorversion\pdfminorversion
    \fi
  \fi
\fi
\usepackage[accsupp]{axessibility}
\ifdefined\XeTeXversion
  
  \let\pdfoptionpdfminorversion\undefined
  
  \let\pdfglyphtounicode\undefined
\fi

\DeclareCaptionStyle{ruled}{labelfont=normalfont,labelsep=colon,strut=off}
\floatstyle{ruled}
\newfloat{listing}{tb}{lst}{}
\floatname{listing}{Listing}
\newsavebox{\SPDocReaderTableBox}
\newcommand{\SPFitToWidth}[2]{%
  \sbox{\SPDocReaderTableBox}{#2}%
  \ifdim\wd\SPDocReaderTableBox>#1\relax
    \resizebox{#1}{!}{\usebox{\SPDocReaderTableBox}}%
  \else
    \usebox{\SPDocReaderTableBox}%
  \fi}
\makeatletter
\def\munderbar#1{\underline{\sbox\tw@{$#1$}\dp\tw@\z@\box\tw@}}
\makeatother

\AddToHook{cmd/appendix/before}{%
  \setcounter{axiom}{0}%
}

\newcommand{\be}{\begin{equation}}
\newcommand{\ee}{\end{equation}}
\newcommand{\bea}{\begin{equation*}\begin{aligned}}
\newcommand{\eea}{\end{aligned}\end{equation*}}

\title{SP-DocReader: Difference-Aware Self-Play for Precise Document OCR}
\runningtitle{SP-DocReader: Difference-Aware Self-Play for Precise Document OCR}

\makeatletter
\renewcommand\AB@affilsepx{\protect\\\protect\Affilfont}
\makeatother

\author[1]{Wenjie Liao}
\author[1]{Xiaohui Song}
\author[2]{Liangjie Zhao}
\author[1]{Haonan Lu}
\affil[1]{Guangdong OPPO Mobile Telecommunications Corp.,Ltd\protect\newline
{\footnotesize\href{mailto:jie3040@akane.waseda.jp}{\texttt{jie3040@akane.waseda.jp}},
\href{mailto:songxiaohui@oppo.com}{\texttt{songxiaohui@oppo.com}},
\href{mailto:luhaonan@oppo.com}{\texttt{luhaonan@oppo.com}}}}
\affil[2]{Adelaide University\protect\newline
{\footnotesize\href{mailto:zhaoliangjie55@gmail.com}{\texttt{zhaoliangjie55@gmail.com}}}}

\hypersetup{
  pdftitle={SP-DocReader: Difference-Aware Self-Play for Precise Document OCR},
  pdfauthor={Wenjie Liao, Xiaohui Song, Liangjie Zhao, Haonan Lu},
  colorlinks=true,
  linkcolor=blue!50!black,
  citecolor=blue!50!black,
  urlcolor=blue!50!black
}

\begin{document}
\raggedbottom

\begin{abstract}
Accurate page transcription remains difficult for vision language models under limited input and training budgets. We present SP-DocReader, a self-play framework for optical character recognition (OCR) that targets residual errors after supervised fine-tuning. Reading Discrepancy Masking aligns reference and generated model tokens through a longest common subsequence, then scores unmatched positions with their full conditioning prefixes. Focused Fidelity Loss adds direct negative log-likelihood supervision at unmatched ground-truth positions. Only the OCR module is trained, while the backbone remains frozen. We derive the combined gradient to distinguish relative score optimization from direct supervision. Compared with SFT-2, SP-DR-3 reduces Vary-600K character error rate on both backbones. On Qwen3-VL-4B, it reduces character error rate by approximately 54 percent and improves DocVQA Average Normalized Levenshtein Similarity (ANLS) by 3.7 points. These results show the value of focusing self-play training on the discrepancies that remain after supervised fine-tuning.
\end{abstract}
\maketitle

\section{Introduction}
\label{sec:intro}

Page-level transcription must reproduce text across a document, including small characters and numbers. DocFormer and LayoutLM represent text and spatial information \cite{appalaraju2021docformer,xu2020layoutlm}. Document images and infographics combine dense text with varied layouts \cite{mathew2021docvqa,mathew2022infographicvqa}. Recent vision language models (VLMs) support document understanding and multimodal reasoning \cite{bai2025qwen3vl,chen2025internvl25,luo2024monointernvl}, but page transcription differs from producing a short answer to a document question.

Higher resolution can improve small-text reading while increasing visual tokens, as in PaliGemma \cite{beyer2024paligemma}. InternLM-XComposer-2.5 uses image tiles \cite{zhang2024internlm}, while LLaVA-OneVision combines AnyRes processing with pooling to control token budgets \cite{li2024llava}. ScreenAI studies optical character recognition (OCR) text as an additional input \cite{baechler2024screenai}, and LATIN-Prompt preserves layout through spaces and line breaks \cite{wang2023layout}. DocVLM compresses OCR text and layout into queries for a frozen VLM \cite{nacson2025docvlm}. We study reduced-resolution page transcription, distinguishing source resolution from the visual tokens passed to the language model.

Our study starts from a reader adapted by supervised fine-tuning (SFT) on annotated pages. Full-sequence SFT supervises every target position without selecting positions from discrepancies in a generated reading. SP-DocReader uses these discrepancies to focus self-play training on the tokens that still need correction.

SP-DocReader builds on DocVLM's modular reader \cite{nacson2025docvlm} and follows SPIN's iterative pairing of references with previous-model responses \cite{chen2024self}. Reading Discrepancy Masking (RDM) aligns the model-token sequences through a longest common subsequence and selects unmatched positions in both readings. The contrastive loss compares score changes against the previous model at these positions while retaining each sequence's full conditioning prefixes. Focused Fidelity Loss (FFL) adds direct negative log-likelihood supervision at unmatched ground-truth positions. Only the OCR module is updated, with the visual encoder and language backbone frozen.

Compared with SFT-2, SP-DocReader reduces transcription error on both backbones and improves most evaluated question-answering scores. On Qwen3-VL-4B, the gains extend from in-distribution reading to zero-shot document benchmarks.

\begin{samepage}
We make three main contributions.

\begin{itemize}
\item We formulate a discrepancy-based self-play objective for page transcription that combines masked relative scoring with direct supervision at unmatched ground-truth positions.
\item We derive the combined gradient to explain the roles of relative and direct supervision, and clarify why selecting loss positions still requires full-prefix computation.
\item We evaluate SP-DocReader on bilingual transcription and zero-shot document benchmarks, and analyze its components, resolution robustness and training cost.
\end{itemize}
\end{samepage}

\section{Related Work}
\label{sec:related}

\noindent\textbf{Efficient Document Reading with VLMs.} Document readers combine image features, recognized text and layout in different ways. ScreenAI uses OCR text as an additional input \cite{baechler2024screenai}, DocLLM models text and boxes without an image encoder \cite{wang2024docllm}, and Hi-VT5 summarizes pages for multi-page question answering \cite{tito2023hierarchical}. DocVLM uses a DocFormerV2-based OCR encoder \cite{nacson2025docvlm,appalaraju2024docformerv2} and compresses its outputs into learnable queries for a frozen VLM \cite{nacson2025docvlm}. Building on this reader, we introduce discrepancy-focused self-play training for page transcription.

\vspace{\baselineskip}
\noindent\textbf{Specialized OCR Systems for Document Recognition.} PP-OCR integrates text detection, direction classification and recognition \cite{du2020ppocr}. Nougat produces markup from scientific document images \cite{blecher2023nougat}, while GOT-OCR2.0 supports plain and formatted OCR \cite{wei2024gotocr}. MinerU combines layout analysis and specialized recognition for document parsing \cite{wang2024mineru}. DeepSeek-OCR studies optical compression of text into visual representations \cite{wei2025deepseekocr}. These systems emphasize recognition and parsing designs. We focus on discrepancy-based training of a modular reader.

\vspace{\baselineskip}
\noindent\textbf{Supervised and Self-Play Fine-Tuning.} For document recognition, we use paired images and transcriptions for SFT. Given input images $I$ drawn from a distribution $q(\cdot)$ and ground-truth transcriptions $y$ following $p_{data}(\cdot\mid I)$, SFT minimizes the expected negative log-likelihood,
\begin{equation}\label{eqn-sft-vlm}
\mathcal{L}_{SFT}(\theta)=-\mathbb{E}_{I\sim q(\cdot),\, y\sim p_{data}(\cdot\mid I)}[\log p_{\theta}(y\mid I)].
\end{equation}
By autoregressive factorization, SFT sums token losses across the reference reading. SP-DocReader uses the model's generated reading to select positions for self-play training.

Direct Preference Optimization (DPO) trains on preferred and dispreferred responses without fitting a separate reward model \cite{rafailov2023direct}. SPIN pairs human demonstrations with responses from the previous model and optimizes a full-response relative score \cite{chen2024self}. SPACE uses separate classification terms for real and generated responses, with full-response scores relative to an opponent \cite{wang2025space}. SPPO frames preference alignment as a constant-sum game \cite{wu2024selfplaypreferenceoptimizationlanguage}, while DNO improves a policy iteratively using on-policy responses and preference feedback \cite{rosset2024direct}. SP-DocReader instead uses the known page transcription to locate unmatched model-token positions in reference and generated readings. It compares the two readings at those positions and directly supervises unmatched reference tokens, while retaining their original conditioning prefixes. Appendix~\ref{app:related} reviews further work on visual-language alignment, adversarial learning and iterative self-improvement.

\section{Methodology}
\label{sec:method}

\begin{figure*}[t]
    \centering    
    \includegraphics[width=\textwidth]{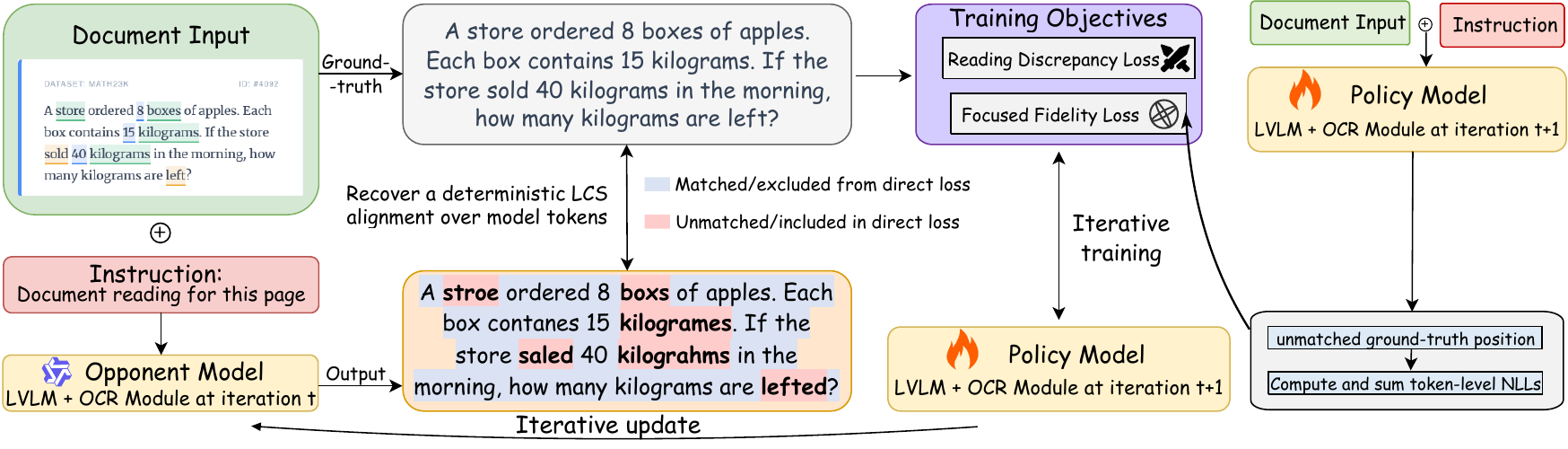}
    \caption{Overview of the SP-DocReader training framework. A frozen opponent generates a synthetic reading of the document, and Reading Discrepancy Masking aligns it with the ground truth to isolate the errors. The main player then optimizes a dual objective that combines the Reading Discrepancy Loss for difference-aware correction with the Focused Fidelity Loss for direct supervision on the hard tokens, after which the opponent is refreshed.}
    \label{fig:2}
\end{figure*}

SP-DocReader adapts the modular reader of DocVLM \cite{nacson2025docvlm} for page transcription. Reading Discrepancy Masking (RDM) selects positions for the Reading Discrepancy Loss, our contrastive objective. Focused Fidelity Loss (FFL) adds direct supervision at unmatched ground-truth positions. Both losses score tokens with their original prefixes.

\subsection{Problem Formulation and Document Reading}

We model page transcription as conditional sequence generation. The input is $(I,x)$, where $I$ is a document image and $x$ is a natural language instruction. The reference sequence $y=(y_1,\dots,y_T)$ contains tokens from a fixed vocabulary $\mathcal{V}$. The reference $y$ transcribes $I$, whereas a generated reading may not cover the page.

We follow the modular reader design of DocVLM \cite{nacson2025docvlm}. The visual encoder maps $I$ to features $\Phi(I)$, while a fixed external OCR engine extracts text and bounding boxes. The OCR branch encodes this text and layout and uses learnable queries to produce compact representations $Q_{\mathrm{ocr},\theta}(I)$. The parameters $\theta$ belong to the OCR encoder, projection layer and learnable queries. The visual encoder and language backbone remain frozen. The reference likelihood factorizes autoregressively as
\begin{equation}\label{eq:factorization}
\begin{aligned}
P_{\theta}(y\mid I,x)={}&\prod_{k=1}^{T}P_{\theta}\!\bigl(y_k\mid y_{<k},\\
&\Phi(I),Q_{\mathrm{ocr},\theta}(I),x\bigr),
\end{aligned}
\end{equation}
where $y_{<k}$ is the prefix preceding position $k$. Standard supervised fine-tuning maximizes the ground-truth likelihood under \eqref{eq:factorization}. Self-play uses the unmatched positions defined below to select direct loss terms. Matched tokens remain in the conditioning prefixes.

\subsection{Iterative Self-Play Training}

Self-play starts from the OCR parameters $\theta_0$ obtained by supervised fine-tuning on the training set $\mathcal{D}$. At round $t$, the model $P_{\theta_t}$ is held fixed and serves as the opponent. It generates a reading $\hat{y}=\operatorname{Gen}_{\theta_t}(I,x)$, where $\operatorname{Gen}$ denotes the decoding procedure. The current model $P_{\theta}$, called the main player, is updated using pairs of $y$ and $\hat{y}$. Their discrepancies determine the loss positions. Appendix~\ref{app:implementation} gives the decoding configuration.

\vspace{\baselineskip}
\noindent \textbf{Reading Discrepancy Masking (RDM).} RDM aligns the generated reading $\hat{y}$ with the ground truth $y$ to identify unmatched positions. It computes the Longest Common Subsequence (LCS) directly over the model token sequences $y$ and $\hat{y}$. Tokens in a subsequence retain their order but need not be adjacent. Let $\mathcal{S}_{common}$ be the set of ordered token subsequences shared by these sequences. The alignment operator selects a longest common subsequence,
\begin{equation}\label{eq:lcs}
\psi(y,\hat{y})\in\operatorname*{arg\,max}_{s\in\mathcal{S}_{common}}|s|.
\end{equation}
We recover its matched positions from the sequence starts using dynamic-programming lengths for all suffix pairs. Equal tokens are matched and both indices advance. Otherwise, we advance the index that preserves the larger suffix LCS length, advancing only the synthetic index on ties. Appendix~\ref{app:implementation} gives the recurrence in \eqref{eq:lcs_suffix} and details EOS handling. Let $\mathcal{E}_y$ and $\mathcal{E}_{\hat{y}}$ be the unmatched positions in the original ground-truth and synthetic sequences. Selecting these positions in increasing order gives
\begin{equation}\label{eq:diff}
y^{\prime}=(y_k)_{k\in\mathcal{E}_y},\qquad \hat{y}^{\prime}=(\hat{y}_k)_{k\in\mathcal{E}_{\hat{y}}}.
\end{equation}
The end-of-sequence (EOS) token follows the same rule. These sets select the tokens to score without removing matched tokens from their conditioning prefixes. For either full sequence $z\in\{y,\hat{y}\}$ and its position set $\mathcal{E}_z$, we define the masked log-likelihood score
\begin{equation}\label{eq:masked_score}
S_{\theta}(z,\mathcal{E}_z\mid I,x)
=\sum_{k\in\mathcal{E}_z}\log P_{\theta}(z_k\mid z_{<k},I,x).
\end{equation}
Conditioning on $(I,x)$ abbreviates the multimodal inputs in \eqref{eq:factorization}. Ground-truth tokens use the full prefix $y_{<k}$, whereas synthetic tokens use their own full prefix $\hat{y}_{<k}$. The current model and fixed opponent score the same tokens at the same positions with the same prefixes, but each uses its own OCR representations. We sum the selected log probabilities without length normalization. The shortened sequences $y^{\prime}$ and $\hat{y}^{\prime}$ are not fed back as new sequences for scoring, and $\exp(S_{\theta})$ is not treated as a normalized probability over these shortened sequences.

\vspace{\baselineskip}
\noindent \textbf{Contrastive Score Optimization.} We adapt SPIN's contrastive score formulation \cite{chen2024self} to compare reference and generated readings at their unmatched positions. The comparison is motivated by the expectation gap used in Integral Probability Metrics (IPMs) \cite{muller1997integral}. Let $\mathcal{F}_t$ be a class of scoring functions. Each function assigns a scalar score to a full reading at selected positions with its original context. We leave the instruction $x$ implicit in $f\in\mathcal{F}_t$. The objective favors a higher score for the ground truth than for the generated reading,
\begin{equation}\label{eq:ipm}
\max_{f\in\mathcal{F}_t}\ \mathbb{E}_{\mathcal{D}}\!\left[f(I,y,\mathcal{E}_y)-f(I,\hat{y},\mathcal{E}_{\hat{y}})\right],
\end{equation}
where $\mathbb{E}_{\mathcal{D}}$ averages over $(I,x,y)$ in the training set, with $\hat{y}=\operatorname{Gen}_{\theta_t}(I,x)$ fixed within the round. We use a monotonically decreasing convex loss $\ell$ to penalize a small score gap,
\begin{equation}\label{eq:surrogate}
\min_{f\in\mathcal{F}_t}\ \mathbb{E}_{\mathcal{D}}\!\left[\ell\!\left(f(I,y,\mathcal{E}_y)-f(I,\hat{y},\mathcal{E}_{\hat{y}})\right)\right].
\end{equation}
To motivate the comparison with the opponent, consider a fixed complete-sequence scorer $g(I,y)$, with $x$ implicit. For fixed $(I,x)$ and $\beta>0$, assume the reference distribution is normalized over complete sequences and the normalizing constant below is finite. Over distributions supported by the reference, the well-defined objective $\mathbb{E}_{y\sim P}[g(I,y)]-\beta D_{KL}(P\,\|\,P_{\theta_t})$ with a Kullback-Leibler penalty \cite{kullback1951information} has the following complete-sequence optimum, as used in DPO \cite{rafailov2023direct}
\begin{equation}\label{eq:optimal}
P^{*}(y\mid I,x)\propto P_{\theta_t}(y\mid I,x)\exp\!\left(\tfrac{1}{\beta}g(I,y)\right).
\end{equation}
If this optimum is realizable as $P_{\theta}$, then $g(I,y)=\beta\log\big(P_{\theta}(y\mid I,x)/P_{\theta_t}(y\mid I,x)\big)$ up to an input-dependent constant. This complete-sequence relation motivates our opponent-relative score. RDM applies the tokenwise comparison at selected original positions. For the current model $\theta$ and fixed opponent $\theta_t$, define
\begin{equation}\label{eq:masked_ratio}
\begin{aligned}
\Delta_t(z)={}&S_{\theta}(z,\mathcal{E}_z\mid I,x)\\
&-S_{\theta_t}(z,\mathcal{E}_z\mid I,x).
\end{aligned}
\end{equation}
Both terms use the same full sequence $z$ and position set $\mathcal{E}_z$. Setting $f(I,z,\mathcal{E}_z)=\beta\Delta_t(z)$ and using the logistic surrogate $\ell(u)=\log(1+e^{-u})$ in \eqref{eq:surrogate} yields the Reading Discrepancy Loss
\begin{equation}\label{eq:rdm}
\begin{aligned}
&\mathcal{L}_{RDM}(\theta,\theta_t)
= \\
&\mathbb{E}_{\mathcal{D}}\!\Bigl[\log\!\bigl(1+\exp\!\bigl(\beta[\Delta_t(\hat{y})-\Delta_t(y)]\bigr)\bigr)\Bigr].
\end{aligned}
\end{equation}
The loss favors a larger score increase over the opponent for the ground-truth discrepancies than for the synthetic discrepancies.

\vspace{\baselineskip}
\noindent \textbf{Focused Fidelity Loss (FFL).} The relative objective in \eqref{eq:rdm} can decrease by lowering the synthetic score without increasing the ground-truth score. FFL therefore adds direct supervision at the unmatched ground-truth positions. Under teacher forcing, each target token is scored using its ground-truth prefix. FFL sums the negative log-likelihood over the same position set $\mathcal{E}_y$,
\begin{equation}\label{eq:ffl}
\begin{aligned}
&\mathcal{L}_{FFL}(\theta)
=\\
&-\mathbb{E}_{\mathcal{D}}\!\Bigl[\sum_{k\in\mathcal{E}_y}\log P_{\theta}\!\left(y_k\mid y_{<k},I,x\right)\Bigr].
\end{aligned}
\end{equation}
The round's readings and masks remain fixed during optimization. We suppress this round dependence in $\mathcal{L}_{FFL}(\theta)$. Both losses retain the original prefixes, and FFL adds no direct loss at matched positions.\newline

\begin{algorithm}[t]
\caption{Iterative Self-Play Training for SP-DocReader}
\label{alg:sp_docreader}
\begin{algorithmic}[1]
\REQUIRE Training set $\mathcal{D} = \{(I_i,x_i,y_i)\}_{i=1}^{N}$, SFT-initialized OCR parameters $\theta_0$, rounds $R\in\mathbb{Z}_{>0}$, loss coefficients $\beta>0$ and $\lambda>0$.
\FOR{$t = 0$ \textbf{to} $R-1$}
    \STATE Fix opponent $\theta_t$ for this round
    \STATE Initialize training records $\mathcal{B}_t\leftarrow\emptyset$
    \FOR{$i=1$ \textbf{to} $N$}
        \STATE $\hat{y}_i\leftarrow\operatorname{Gen}_{\theta_t}(I_i,x_i)$
        \STATE Recover $\mathcal{E}_{y_i},\mathcal{E}_{\hat{y}_i}$ from the LCS alignment in \eqref{eq:lcs} with the fixed tie rule
        \STATE Add $(I_i,x_i,y_i,\hat{y}_i,\mathcal{E}_{y_i},\mathcal{E}_{\hat{y}_i})$ to $\mathcal{B}_t$
    \ENDFOR
    \STATE $\theta_{t+1}\leftarrow\operatorname{Optimize}(\theta_t,\mathcal{B}_t;\beta,\lambda)$ (Eq.~\eqref{eq:update})
\ENDFOR
\STATE \textbf{Return} OCR parameters $\theta_R$
\end{algorithmic}
\end{algorithm}

\noindent \textbf{Training Objective and Opponent Update.} The training objective adds FFL to the masked contrastive loss,
\begin{equation}\label{eq:total}
\mathcal{L}_{total}(\theta,\theta_t)=\mathcal{L}_{RDM}(\theta,\theta_t)+\lambda\,\mathcal{L}_{FFL}(\theta),
\end{equation}
with $\lambda>0$ controlling the weight of direct supervision. Pairs with only one empty mask are retained. Pairs with both masks empty contribute no loss gradient, but batch averaging still uses the original number of samples, as specified in Appendix~\ref{app:implementation}, \eqref{eq:batch_average}. The round's training records $\mathcal{B}_t$ retain all samples, including their full sequences and masks. We write the update as
\begin{equation}\label{eq:update}
\theta_{t+1}\leftarrow\operatorname{Optimize}(\theta_t,\mathcal{B}_t;\beta,\lambda),
\end{equation}
where $\operatorname{Optimize}$ initializes the trainable OCR parameters at $\theta_t$ and applies AdamW to \eqref{eq:total} under a finite training budget. The opponent and training records remain fixed during this update. The notation does not assume an exact global minimizer. The updated model with OCR parameters $\theta_{t+1}$ becomes the next round's opponent and generates new readings and masks. Algorithm \ref{alg:sp_docreader} summarizes this procedure.

\subsection{Optimization Analysis}

The two losses play complementary roles in each update. The Reading Discrepancy Loss compares how much the current model improves on the opponent's scores for reference and generated readings. When the reference has a small relative advantage, the comparison gives more weight to raising its score and lowering the generated score. FFL adds direct pressure to raise the reference score, so correction does not depend only on penalizing the generated reading. Equation~\eqref{eq:masked_score} scores unmatched tokens using their original prefixes, with matched tokens still supplying context. The loss therefore targets reading disagreements without treating the extracted fragments as new sequences. Appendix~\ref{app:derivations} derives these contributions in \eqref{eq:pair_gradient}.

\section{Experiments}
\label{sec:experiments}

\subsection{Experimental Setup}

\begin{table*}[t]
\caption{Reading accuracy and zero-shot generalization. Vary-600K is in-distribution while DocBank, IIT-CDIP, DocVQA and InfographicVQA are zero-shot. SFT-2 is trained for two supervised epochs and SP-DR-3 follows three self-play iterations. Parentheses show SP-DR-3's change over SFT-2, with improvements in red and regressions in blue. CER is reported as a percentage. Full metrics and iteration trajectories appear in Appendix~\ref{app:trajectories}.}
\centering
\footnotesize
\setlength{\tabcolsep}{2pt}
\SPFitToWidth{\textwidth}{%
\begin{tabular}{@{\extracolsep{\fill}}cccccccccc}
\toprule
\multirow{2}{*}{Models} & \multirow{2}{*}{Method} & \multicolumn{2}{c}{Vary-600K} & \multicolumn{2}{c}{DocBank} & \multicolumn{2}{c}{IIT-CDIP} & \multicolumn{2}{c}{VQA (ANLS$\uparrow$)} \\
\cmidrule(lr){3-4} \cmidrule(lr){5-6} \cmidrule(lr){7-8} \cmidrule(lr){9-10}
& & CER$\downarrow$ & NED$\uparrow$ & CER$\downarrow$ & NED$\uparrow$ & CER$\downarrow$ & NED$\uparrow$ & DocVQA & InfoVQA \\
\midrule
\multirow{3}{*}{\begin{tabular}[c]{@{}c@{}}InternVL\\3.5-4B\end{tabular}}
& Base (low-res) & 4.85 & 0.938 & 13.44 & 0.812 & 19.56 & 0.735 & 72.79 & 50.52 \\
& SFT-2 & 2.12 & 0.971 & 11.60 & 0.848 & 17.83 & 0.778 & 73.01 & 48.15 \\
& SP-DR-3 & 1.17{\scriptsize (\textcolor{red}{-0.95})} & 0.988{\scriptsize (\textcolor{red}{+0.017})} & 9.73{\scriptsize (\textcolor{red}{-1.87})} & 0.884{\scriptsize (\textcolor{red}{+0.036})} & 15.35{\scriptsize (\textcolor{red}{-2.48})} & 0.823{\scriptsize (\textcolor{red}{+0.045})} & 74.27{\scriptsize (\textcolor{red}{+1.26})} & 48.06{\scriptsize (\textcolor{blue}{-0.09})} \\
\midrule
\multirow{3}{*}{\begin{tabular}[c]{@{}c@{}}Qwen3\\VL-4B\end{tabular}}
& Base (low-res) & 5.22 & 0.931 & 12.82 & 0.834 & 18.45 & 0.756 & 78.54 & 52.21 \\
& SFT-2 & 2.40 & 0.967 & 10.95 & 0.868 & 16.71 & 0.802 & 81.22 & 55.34 \\
& SP-DR-3 & 1.11{\scriptsize (\textcolor{red}{-1.29})} & 0.987{\scriptsize (\textcolor{red}{+0.020})} & 9.05{\scriptsize (\textcolor{red}{-1.90})} & 0.898{\scriptsize (\textcolor{red}{+0.030})} & 14.42{\scriptsize (\textcolor{red}{-2.29})} & 0.847{\scriptsize (\textcolor{red}{+0.045})} & 84.89{\scriptsize (\textcolor{red}{+3.67})} & 58.94{\scriptsize (\textcolor{red}{+3.60})} \\
\bottomrule
\end{tabular}
}
\label{tab:main}
\end{table*}

\begin{figure*}[t]
  \centering
  \includegraphics[width=\textwidth]{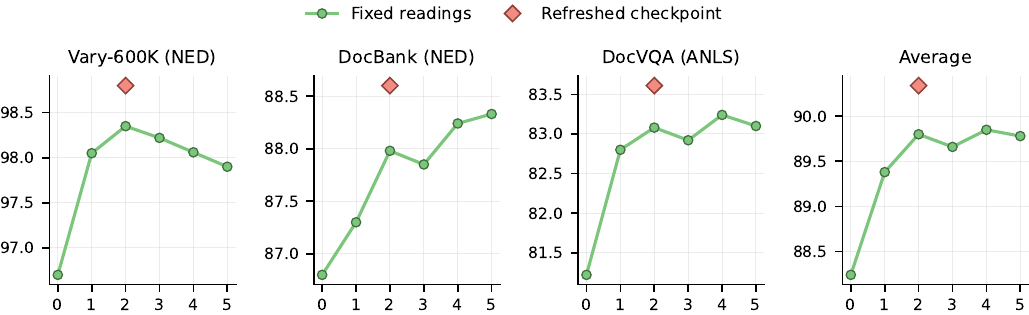}
  \caption{Fixed and refreshed self-play training on Qwen3-VL-4B. Green circles show the fixed-reading branch from its starting checkpoint through five local epochs. The red diamond represents a single checkpoint after two local epochs in the refreshed second round, which updates the opponent, generated readings, and discrepancy masks. The panels report Vary-600K NED, DocBank NED, DocVQA ANLS, and their mean, with NED scaled to 0--100 for the mean. The refreshed checkpoint scores higher than the fifth fixed-reading epoch in every panel.}
  \label{fig:epochs}
\end{figure*}

\noindent \textbf{Base Models and Architecture.} We integrate SP-DocReader with two vision language models, InternVL-3.5-4B \cite{internvl3_5} and Qwen3-VL-4B \cite{bai2025qwen3vl}. Following DocVLM \cite{nacson2025docvlm}, we attach a specialized OCR modality to each backbone. EasyOCR \cite{easyocr} supplies recognized text and bounding boxes and is not updated during training. DocVLM's OCR encoder is derived from the T5-based DocFormerV2 model \cite{appalaraju2024docformerv2,t5_raffel}. We adapt this encoder's input representation using the layout and text interleaving approach of LayTextLLM \cite{laytextllm}. Appendix~\ref{app:implementation} specifies this adaptation. For the main SP-DocReader model, the visual encoder and language backbone remain frozen, and only the OCR encoder, projection layer and learnable queries are updated.

\vspace{\baselineskip}
\noindent \textbf{Datasets and Training Protocol.} We draw training pages from Vary-600K \cite{vary_dataset}. We use a balanced subset of 30{,}000 English and 30{,}000 Chinese pages for the supervised and self-play stages. Training uses two supervised epochs followed by self-play with generated readings. We use SFT-2 to denote the checkpoint after two supervised epochs and SP-DR-$r$ to denote the checkpoint after $r$ self-play rounds initialized from SFT-2. A self-play round is distinct from an epoch, and SP-DR-3 is the default model reported in the main results. Optimization settings and experimental protocols appear in Appendix~\ref{app:implementation}.

\vspace{\baselineskip}
\noindent \textbf{Evaluation Benchmarks and Metrics.} We measure reading accuracy on the Vary-600K validation set and test zero-shot generalization on DocBank \cite{docbank} and IIT-CDIP \cite{iit_cdip}, which cover academic pages and scanned documents, respectively. We also evaluate question answering on DocVQA \cite{mathew2021docvqa} and InfographicVQA \cite{mathew2022infographicvqa}. Additional evaluations use OmniDocBench \cite{ouyang2024omnidocbench}, OCRBench v2 \cite{fu2024ocrbenchv2} and CC-OCR \cite{yang2024ccocr}. The experiments use reduced-resolution inputs, with PDF pages rendered at 72 DPI and raster-image preprocessing specified in Appendix~\ref{app:implementation}. We report Character Error Rate (CER) as a percentage and retain the higher-is-better edit score labeled NED in the tables. Appendix~\ref{app:implementation} defines their computation from character edit distance \cite{levenshtein1966distance}. Question answering uses Average Normalized Levenshtein Similarity (ANLS) \cite{biten2019scene} on a 0 to 100 scale. Appendix~\ref{app:benchmarks} describes the datasets and data audit, reference checks and benchmark scoring. Here, zero-shot refers to evaluation without target-benchmark adaptation during the supervised and self-play stages described in this paper. It does not imply that the pretrained backbones or OCR components have never encountered these datasets or related documents.

\begin{figure*}[t]
  \centering
  \includegraphics[width=\textwidth]{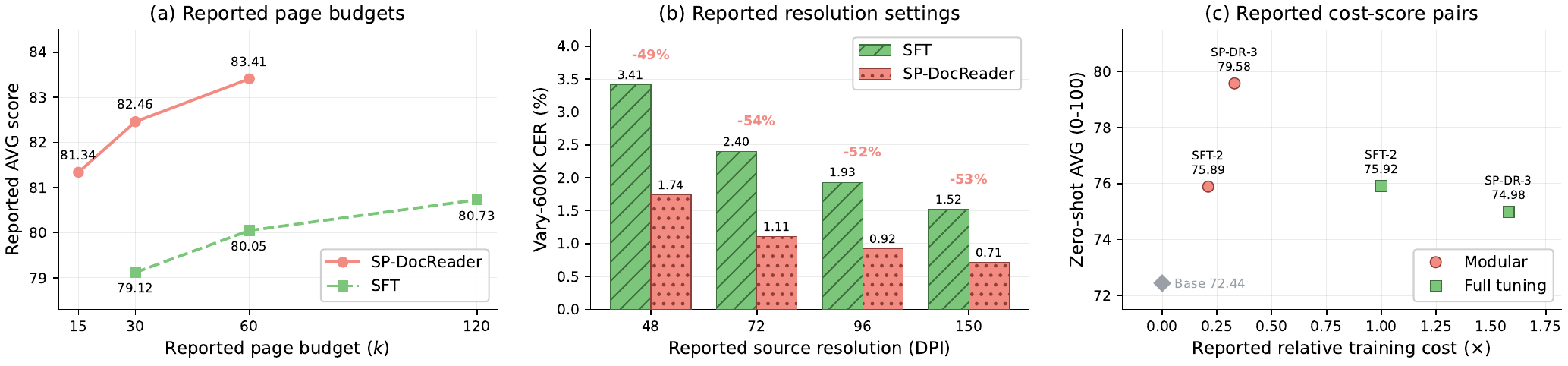}
  \caption{Reported Qwen3-VL-4B comparisons across training-page budgets, source-image resolutions and relative training costs. Panel (a) averages Vary-600K, DocBank and IIT-CDIP NED with DocVQA and InfographicVQA ANLS, weighting each dataset equally and scaling NED by 100. Panel (b) compares CER at four resolutions. Panel (c) uses the mean of DocBank, IIT-CDIP NED, DocVQA and InfographicVQA ANLS.}
  \label{fig:analysis}
\end{figure*}

\subsection{Main Results}

\noindent \textbf{Recognition and Generalization.} Table~\ref{tab:main} compares SP-DR-3 with SFT-2 on transcription and question answering. Vary-600K CER falls from 2.40 to 1.11 on Qwen3-VL-4B and from 2.12 to 1.17 on InternVL-3.5-4B. On Qwen3-VL-4B, zero-shot CER falls by 1.90 percentage points on DocBank and 2.29 on IIT-CDIP, while ANLS rises by 3.67 points on DocVQA and 3.60 on InfographicVQA. On InternVL-3.5-4B, transcription improves across all three reading benchmarks, while InfographicVQA ANLS changes slightly from 48.15 to 48.06. The transcription gains thus extend from the training domain to both zero-shot reading datasets. DocVQA also improves on both backbones, showing that better page reading accompanies stronger question answering in this benchmark. Appendix~\ref{app:trajectories} reports the full supervised and self-play trajectories and specifies the metric sets used in the reported averages.

\vspace{\baselineskip}
\noindent \textbf{Results on Additional Benchmarks.} Table~\ref{tab:ocrcompare} reports aggregate scores on OmniDocBench, OCRBench v2 and CC-OCR. SP-DR-3 improves all three aggregate scores over SFT-2 on both backbones. On Qwen3-VL-4B, OmniDocBench edit distance decreases from 0.347 to 0.291, while OCRBench v2 rises from 49.1 to 55.3 and CC-OCR from 65.4 to 71.0. These results extend the gains beyond the five primary evaluation datasets.\newline

\begin{table}[t]
\caption{Results on three additional document parsing benchmarks. Lower is better for the OmniDocBench edit distance and higher is better for the other two.}
\centering
\footnotesize
\setlength{\tabcolsep}{4pt}
\SPFitToWidth{\columnwidth}{%
\begin{tabular}{llccc}
\toprule
Models & Method & \begin{tabular}[c]{@{}c@{}}OmniDoc\\Edit$\downarrow$\end{tabular} & \begin{tabular}[c]{@{}c@{}}OCRBench\\v2$\uparrow$\end{tabular} & \begin{tabular}[c]{@{}c@{}}CC-OCR\\$\uparrow$\end{tabular} \\
\midrule
\multirow{3}{*}{\begin{tabular}[c]{@{}c@{}}InternVL\\3.5-4B\end{tabular}} & Base (low-res) & 0.439 & 40.1 & 55.8 \\
 & SFT-2 & 0.366 & 46.5 & 62.7 \\
 & SP-DR-3 & 0.312{\scriptsize (\textcolor{red}{-0.054})} & 52.0{\scriptsize (\textcolor{red}{+5.5})} & 67.9{\scriptsize (\textcolor{red}{+5.2})} \\
\midrule
\multirow{3}{*}{\begin{tabular}[c]{@{}c@{}}Qwen3\\VL-4B\end{tabular}} & Base (low-res) & 0.418 & 42.6 & 58.3 \\
 & SFT-2 & 0.347 & 49.1 & 65.4 \\
 & SP-DR-3 & 0.291{\scriptsize (\textcolor{red}{-0.056})} & 55.3{\scriptsize (\textcolor{red}{+6.2})} & 71.0{\scriptsize (\textcolor{red}{+5.6})} \\
\bottomrule
\end{tabular}
}
\label{tab:ocrcompare}
\end{table}

\noindent \textbf{Fixed versus refreshed training records.} Figure~\ref{fig:epochs} compares continued training on fixed readings with a refreshed second-round checkpoint on Qwen3-VL-4B. The refreshed checkpoint reaches an average near 90.3 after two local epochs, above the fixed-reading branch's average near 89.8 after five local epochs. Each plotted metric also favors the refreshed checkpoint over the fifth fixed-reading epoch. Refreshing updates the opponent, generated readings and LCS masks together, giving the next round new errors to learn from. Appendix~\ref{app:implementation} describes the separate refresh control with matched update counts.

\subsection{Ablation Study}

\noindent \textbf{Component Analysis.} Table~\ref{tab:component} compares the full model with two ablation rows on Qwen3-VL-4B. Without the contrastive Reading Discrepancy Loss, the model still applies FFL to unmatched reference tokens, but Vary-600K CER rises from 1.11 to 1.88. Without FFL, the model retains the reference-versus-generated comparison, but DocVQA ANLS falls from 84.89 to 82.96. Both reduced variants improve on SFT-2 across all five columns, and combining the losses gives the best score in every column. Appendix~\ref{app:implementation} details the component settings and additional controls.

\vspace{\baselineskip}
\noindent \textbf{Data efficiency.} Figure~\ref{fig:analysis}(a) compares Qwen3-VL-4B results at labeled page budgets of 15k, 30k and 60k for SP-DocReader and 30k, 60k and 120k for SFT. AVG gives equal weight to Vary-600K, DocBank and IIT-CDIP NED, scaled by 100, and DocVQA and InfographicVQA ANLS. It combines in-distribution reading with zero-shot performance, unlike the zero-shot-only AVG in panel (c). At 30k training pages, SP-DocReader reaches an AVG of 82.46, exceeding the 80.73 obtained by SFT with 120k pages. The comparison highlights the value of self-play updates when labeled pages are limited, with Vary-600K included in the overall score. Appendix~\ref{app:implementation} details how distinct reference pages are counted.

\begin{table}[t]
\caption{Component analysis on Qwen3-VL-4B. CER is a percentage and the VQA columns report ANLS.}
\centering
\footnotesize
\setlength{\tabcolsep}{2.5pt}
\SPFitToWidth{\columnwidth}{%
\begin{tabular}{lccccc}
\toprule
Method & \begin{tabular}[c]{@{}c@{}}Vary\\CER$\downarrow$\end{tabular} & \begin{tabular}[c]{@{}c@{}}DocBank\\CER$\downarrow$\end{tabular} & \begin{tabular}[c]{@{}c@{}}IIT-CDIP\\CER$\downarrow$\end{tabular} & DocVQA$\uparrow$ & InfoVQA$\uparrow$ \\
\midrule
SFT-2 & 2.40 & 10.95 & 16.71 & 81.22 & 55.34 \\
w/o RDM & 1.88 & 10.31 & 15.36 & 83.47 & 57.81 \\
w/o FFL & 1.57 & 9.88 & 15.04 & 82.96 & 57.29 \\
SP-DR (full) & \textbf{1.11} & \textbf{9.05} & \textbf{14.42} & \textbf{84.89} & \textbf{58.94} \\
\bottomrule
\end{tabular}
}
\label{tab:component}
\end{table}

\begin{table}[t]
\caption{Architecture ablation on Qwen3-VL-4B.}
\centering
\footnotesize
\setlength{\tabcolsep}{2.5pt}
\SPFitToWidth{\columnwidth}{%
\begin{tabular}{lccccc}
\toprule
Method & \begin{tabular}[c]{@{}c@{}}Vary\\CER$\downarrow$\end{tabular} & \begin{tabular}[c]{@{}c@{}}DocBank\\CER$\downarrow$\end{tabular} & \begin{tabular}[c]{@{}c@{}}IIT-CDIP\\CER$\downarrow$\end{tabular} & DocVQA$\uparrow$ & \begin{tabular}[c]{@{}c@{}}Train.\\Params\end{tabular} \\
\midrule
OCR-free & 1.04 & 11.93 & 17.61 & 79.62 & 3.3B \\
Full-parameter & \textbf{0.98} & 11.28 & 16.87 & 80.41 & 3.8B \\
Modular (ours) & 1.11 & \textbf{9.05} & \textbf{14.42} & \textbf{84.89} & \textbf{0.5B} \\
\bottomrule
\end{tabular}
}
\label{tab:arch}
\end{table}

\begin{figure*}[t]
\centering
\includegraphics[width=\textwidth]{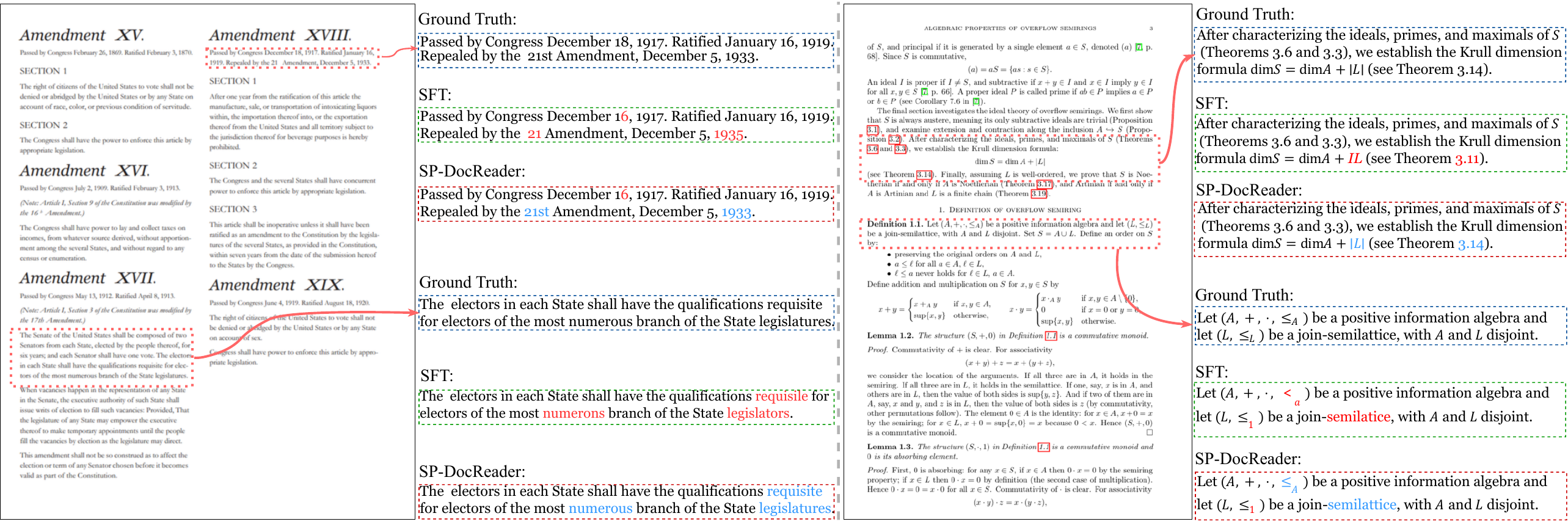}
\caption{Qualitative comparisons on two low-resolution documents. Blue, green and red dashed boxes contain the reference, supervised and SP-DocReader readings, respectively. SP-DocReader corrects the discrepancies.}
\label{fig:qualitative}
\end{figure*}

\vspace{\baselineskip}
\noindent \textbf{Architecture ablations.} Table~\ref{tab:arch} compares adaptation variants on Qwen3-VL-4B. Modular tuning has higher Vary-600K CER than OCR-free and full-parameter tuning, but achieves the lowest CER on DocBank and IIT-CDIP and the highest DocVQA ANLS. The contrast is clearest against full tuning: its Vary-600K CER is 0.98, compared with 1.11 for modular tuning, yet its DocBank and IIT-CDIP CER are 11.28 and 16.87, compared with 9.05 and 14.42. Thus, the best in-domain result does not yield the best zero-shot reading scores in this comparison. Modular tuning achieves those scores with 0.5B trainable parameters, compared with 3.3B for OCR-free tuning and 3.8B for full-parameter tuning. Appendix~\ref{app:modality} provides their training trajectories, while Appendix~\ref{app:implementation} specifies the architecture configurations.

\vspace{\baselineskip}

\noindent \textbf{Resolution robustness.} Figure~\ref{fig:analysis}(b) compares Vary-600K CER from 48 to 150 DPI. SP-DocReader has lower CER than SFT at all four tested resolutions. At 48 DPI, CER is 3.41 for SFT and 1.74 for SP-DocReader. At 150 DPI, it is 1.52 and 0.71, respectively. The advantage persists across the tested resolution range. Appendix~\ref{app:implementation} details the branch inputs, token budgets and resolution protocol.

\subsection{Cost Analysis}

\noindent Figure~\ref{fig:analysis}(c) uses the four-task zero-shot AVG from Table~\ref{tab:modality}. From SFT-2 to the third self-play round, it rises from 75.89 to 79.58 for modular tuning, while full tuning changes from 75.92 to 74.98. The base score is 72.44 from Table~\ref{tab:zeroshot_full}. At the third round, modular tuning reaches an AVG of 79.58 at a relative training cost of 0.33, compared with 74.98 at 1.58 for full tuning. Costs are normalized by the complete joint-tuning SFT-2 adaptation run. The reported inference overhead is about ten percent relative to the backbone without OCR. Appendix~\ref{app:implementation} details the cumulative training-cost accounting and end-to-end inference protocol, including external OCR.

\subsection{Qualitative Results}

\noindent Figure~\ref{fig:qualitative} compares readings of a constitutional document and a mathematics page at low resolution. In the constitutional excerpt, SFT-2 misreads the repeal year and changes ``State legislatures'' to ``State legislators.'' SP-DocReader restores the date and the original wording. On the mathematics page, SFT-2 introduces errors into a dimension formula and reads the reference to Theorem 3.14 as Theorem 3.11. SP-DocReader recovers the theorem number and transcribes the formula more accurately. The lemma excerpt further shows the importance of preserving symbols alongside ordinary text. Together, the examples illustrate how discrepancy-focused training improves recognition of both prose and mathematical notation.

\section{Conclusion}
\label{sec:conclusion}

We introduced SP-DocReader, a self-play framework that trains a modular document reader on differences between generated readings and reference text. Reading Discrepancy Masking selects unmatched model-token positions while preserving their original context, and Focused Fidelity Loss directly supervises unmatched reference tokens. The OCR branch is updated while the visual encoder and language backbone remain frozen.

Compared with two-epoch supervised fine-tuning, three self-play rounds reduce Vary-600K CER from 2.40 to 1.11 on Qwen3-VL-4B and from 2.12 to 1.17 on InternVL-3.5-4B. On Qwen3-VL-4B, DocBank and IIT-CDIP CER decrease by 1.90 and 2.29 percentage points, while DocVQA and InfographicVQA ANLS increase by 3.67 and 3.60 points. The full objective also outperforms either reduced-loss variant in the component analysis. At a 30k-page training budget, SP-DocReader reaches a five-task AVG of 82.46, compared with 80.73 for SFT using 120k pages. Together, these results show that focusing self-play on reading discrepancies improves transcription across the evaluated document datasets and supports strong performance with fewer reference pages in the tested budget comparison.

\section*{Limitations}
\label{sec:limitations}

SP-DocReader focuses on single-page transcription rather than multi-page reading or structured extraction. Its fixed generation budget may limit coverage of longer pages. Our experiments use English and Chinese training pages and evaluate two vision language backbones. The results therefore do not establish generalization to other languages or model families.

\section*{Ethical Considerations}
\label{sec:ethics}

Document transcription can expose personal or confidential information, so its use should respect data access conditions and protect sensitive content. Model outputs may contain recognition errors and should be checked against the source documents before use in consequential decisions. Any reuse or release of data and model components should comply with their applicable licenses and permissions.

\bibliographystyle{acl_natbib}
\bibliography{aaai2027}

\clearpage
\appendix

\section{Additional Related Work}
\label{app:related}

\vspace{2mm}

\noindent \textbf{Fine-grained Alignment in Vision Language Models.} CLIP learns visual representations by matching images with their paired text \cite{radford2021clip}. LLaVA connects a vision encoder to a language model and uses visual instruction tuning for multimodal interaction \cite{liu2024llava}. For text-rich images, Qwen-VL increases the input resolution during training and uses a fixed set of visual queries \cite{bai2023qwenvl}, whereas InternVL 1.5 divides images into tiles for dynamic high-resolution processing \cite{chen2024internvl}. mPLUG-DocOwl uses a visual abstractor and document-oriented instruction tuning to connect visual features with a language model \cite{ye2023mplugdocowl}.

\vspace{2mm}

\noindent \textbf{Adversarial Learning and Iterative Self-Improvement.} Generative adversarial networks train a generator and discriminator through a minimax objective \cite{goodfellow2014generative}. Integral probability metrics compare distributions through expectations over a specified function class \cite{muller1997integral}. The Wasserstein generative adversarial network (WGAN) uses a Lipschitz-constrained critic to approximate the Wasserstein distance \cite{arjovsky2017wasserstein}.

STaR repeatedly generates rationales, filters them by answer correctness, and fine-tunes on the retained solutions \cite{zelikman2022star}. The reinforcement learning from human feedback (RLHF) workflow of \citet{dong2024rlhf} trains a proxy reward model on public preference data and uses it to label response pairs for online policy updates.

Studies of instruction tuning examine how supervised data affects different capabilities \cite{dong2024abilities,wang2023far}. Continual fine-tuning studies report forgetting when language models learn successive tasks \cite{luo2023catastrophic}.

\section{Supporting Derivations}
\label{app:derivations}

\noindent \textbf{Gradient of the combined objective.} Fix an input $(I,x)$, a reading pair $(y,\hat{y})$, the opponent $\theta_t$ and both position sets. For the current model, write
\[
\begin{aligned}
s_y&=S_{\theta}(y,\mathcal{E}_y\mid I,x),\\
s_{\hat{y}}&=S_{\theta}(\hat{y},\mathcal{E}_{\hat{y}}\mid I,x),\\
u&=\beta[\Delta_t(y)-\Delta_t(\hat{y})].
\end{aligned}
\]
The per-pair contribution to \eqref{eq:total} is
\[
\mathcal{L}_{\mathrm{pair}}(\theta)
=\log(1+e^{-u})-\lambda s_y.
\]
Define $\sigma(v)=(1+e^{-v})^{-1}$. The logistic term has derivative $-\sigma(-u)$ with respect to $u$. Since the opponent and masks are fixed,
\[
\nabla_{\theta}u
=\beta(\nabla_{\theta}s_y-\nabla_{\theta}s_{\hat{y}}).
\]
Combining the two terms gives
\begin{equation}\label{eq:pair_gradient}
\begin{aligned}
\nabla_{\theta}\mathcal{L}_{\mathrm{pair}}
={}&-[\beta\sigma(-u)+\lambda]\nabla_{\theta}s_y\\
&+\beta\sigma(-u)\nabla_{\theta}s_{\hat{y}}.
\end{aligned}
\end{equation}
Both score gradients retain the full original prefixes and sum over the selected positions. Averaging these contributions follows \eqref{eq:batch_average}, including pairs with empty masks. This identity describes the existing objective.

\noindent \textbf{Scope of the variance interpretation.} Fix the current parameters $\theta$ and opponent $\theta_t$ and consider variation across the round's training examples. Each example retains its fixed reading pair and masks. In \eqref{eq:pair_gradient}, both score gradients and the coefficient $\beta\sigma(-u)$ can vary across examples. The variance of the combined gradient depends on the variances and covariance of the two weighted contributions, with FFL included in the ground-truth contribution. The number of selected positions alone does not determine these quantities. We make no independence or equal-variance assumption about token contributions and derive no variance bound relative to full-sequence training.

\noindent \textbf{Loss masking and computation.} When vocabulary logits are computed at every position before applying the loss masks, those dense projection operations have already been performed. Reducing projection work through masking requires an implementation that skips the excluded projections. Selected losses can depend on preceding hidden states through contextual computation, so a matched position need not have zero gradient in the internal computation. The frozen backbone weights are not optimized, but its derivatives with respect to the OCR representations are still needed to update the OCR branch.

\noindent \textbf{Scope of cost comparisons.} A complete adaptation-cost comparison must include input preprocessing, supervised initialization, generation, LCS alignment, both current-model and opponent scoring, and optimization over all required rounds. A single step or round does not cover this total, and shared operations must not be counted twice. Runtime and memory claims require measurements with a stated scope.

\section{Implementation Details}
\label{app:implementation}

\noindent \textbf{OCR architecture.} We interleave projected bounding-box embeddings with their OCR text embeddings at the input of the DocFormerV2-derived encoder. This replaces the source design's additive fusion of text and spatial features rather than adding a complete LayTextLLM model. We retain DocVLM's query compression design without the source DocFormerV2 visual branch or decoder. The input embedding layers belong to the trainable OCR encoder. The output projection maps query features to the frozen VLM. Before training, we check input compatibility, encoded length and the trainable parameter count.

\noindent We build SP-DocReader on the Accelerate library for distributed training \cite{gugger2022accelerate}. All runs use two NVIDIA H20 GPUs with 96GB each. We set a batch of 2 per device, which gives a global batch of 4. The projection layer and the learnable queries use a learning rate of $10^{-4}$, and the OCR encoder uses $5\times10^{-5}$. These two learning rates match those used in the initial OCR alignment stage of DocVLM \cite{nacson2025docvlm}. We optimize with AdamW under a cosine schedule, and we warm up over the first ten percent of the steps. The weight $\lambda$ on the Focused Fidelity Loss is set to $0.5$ by default.

The reported training uses two supervised epochs followed by up to five self-play rounds, with the third round as the default model unless stated otherwise. The Vary-600K training subset contains 30{,}000 English and 30{,}000 Chinese pages.

\noindent \textbf{Supervision-budget accounting.} We count distinct reference pages used in project-specific initialization, supervised training and all self-play rounds. Other task-specific training stages also enter this count. Repeated use of the same pages does not create new reference pages. We record references used for development and model selection, their union with the training pages, and the held-out evaluation data separately from the provenance of common pretrained components. Training-reference counts do not measure human annotation time or cost.

\noindent \textbf{Data-efficiency comparison.} For Qwen3-VL-4B, we use training subsets of 15k, 30k, 60k and 120k pages, nested across budgets and balanced between English and Chinese. Each budget starts from the same pre-task initialization and completes its own two-epoch supervised stage. The 15k, 30k and 60k variants then use only their respective training pages for three self-play rounds. A smaller-budget variant does not inherit a checkpoint adapted on a larger training set. Any project-specific initialization training must remain within the stated budget or be included in the reported training union.

\noindent We compare OCR-only modular SFT and self-play under the same input and scoring protocol. The primary comparison uses the 30k SP-DR-3 and 120k SFT-2 checkpoints, with development data recorded separately. We use fixed stage endpoints and the same held-out evaluation sets, while recording cumulative computation separately from reference-page counts.

\noindent \textbf{Reading generation.} We greedily generate one reading per page at each round's start and reuse it within the round. Decoding stops at EOS or 1{,}024 new tokens, excluding the prompt, with no forced EOS, resampling, candidate filtering, or repetition constraints.

\noindent \textbf{Length protocol.} We use a 1{,}024-new-token response budget for self-play generation and evaluation. This budget excludes the prompt and does not cap the multimodal inputs or the teacher-forced reference. We use one decoding call per task request, without continuation or stitching multiple readings. SFT and self-play scoring retain the complete reference and its original prefixes. Each response sequence together with its conditioning input must fit the model and runtime limits. We check sequence lengths before running and do not silently truncate or remove samples.

\noindent \textbf{Coverage evaluation.} We score each capped page transcription against the complete frozen reference, without shortening the reference to the output length or excluding pages because generation reaches the cap. We record reference lengths with each backbone's tokenizer, output lengths and termination reasons. Emitting EOS does not guarantee complete transcription, and reaching the cap alone does not prove an omission. VQA and other tasks retain their task-specific references and evaluation rules.

\noindent \textbf{Transcription scoring.} We fix each reference's reading order and serialization. We apply Unicode NFC normalization and collapse whitespace in both references and predictions, while preserving case, punctuation and content order. CER is the total character edit count divided by the total reference characters, multiplied by 100. For each page, we compute normalized edit similarity as $1-d(r,p)/\max(|r|,|p|)$, where $r$ and $p$ are the normalized reference and prediction and $d$ is character edit distance. We average this similarity over pages, assigning 1 when both strings are empty.

\noindent \textbf{Auxiliary metrics.} We use a fixed SacreBLEU zh tokenizer \cite{post2018sacrebleu} for word error rate (WER), Token-F1, ROUGE-L \cite{lin2004rouge} and BLEU-4 \cite{papineni2002bleu}. Its units include Chinese characters and non-Chinese words or punctuation, rather than model tokens. WER uses corpus edit counts. Token-F1 and ROUGE-L F1 are averaged over pages, and BLEU-4 uses corpus statistics. This evaluation tokenization does not change the model-token LCS or EOS rules used in training.

\noindent \textbf{Branch inputs.} We provide the same reduced-resolution page to the visual preprocessing pipeline and EasyOCR. We render PDFs at 72 DPI when their physical page dimensions are available. For raster-only pages, we preserve aspect ratio and reduce the longer edge to at most 842 pixels, without upsampling smaller sources or assigning an assumed DPI. EasyOCR uses a fixed configuration with a canvas size of 2{,}560 and a magnification ratio of 1.0. Both branches receive the same page content, without a higher-resolution OCR input.

\noindent \textbf{Visual and OCR budgets.} We use one full-page $448\times448$ tile without an extra thumbnail for standard InternVL3.5. For Qwen3-VL, we use an image-pixel budget from $64\times32^2$ to $256\times32^2$, using its native processor. The OCR branch produces 64 query representations per page. We record processed image sizes, tile counts and the actual visual positions passed to the language model. The query count does not limit the OCR encoder's input text.

\noindent \textbf{OCR input checks.} We retain the complete extracted OCR text and boxes, record their encoded length and verify that they fit the OCR encoder before running. Inputs are checked against the encoder capacity without silent truncation or removal. Fixed external OCR outputs can be reused across matched comparisons, but trainable OCR representations must be recomputed. OCR preprocessing and encoding remain part of the system cost.

\noindent \textbf{Resolution check.} We evaluate fixed SFT-2 and SP-DR-3 checkpoints under four source-image settings, without retraining at each setting. For PDFs with known page dimensions, we use 48, 72, 96 and 150 DPI. For raster-only pages, we use long-edge caps of 561, 842, 1{,}123 and 1{,}754 pixels without upsampling. Both branches receive the image for that setting, and OCR outputs are recomputed and shared between the two methods. Visual preprocessing rules, budget limits, query counts and decoding settings remain fixed, with actual token counts recorded. This evaluates the combined response of both input branches to downsampling, rather than isolating the visual encoder.

\noindent \textbf{Component comparisons.} We initialize all component variants from the same modular SFT-2 checkpoint and retain the LCS masks. The full model uses $\mathcal{L}_{\mathrm{RDM}}+0.5\mathcal{L}_{\mathrm{FFL}}$. Removing the contrastive term leaves $0.5\mathcal{L}_{\mathrm{FFL}}$, whereas removing FFL leaves $\mathcal{L}_{\mathrm{RDM}}$. Each variant generates readings with its own previous-round model and refreshes them between rounds. We use two epochs per round and report the third round under the same data and update budget.

\noindent \textbf{Refresh comparison.} For Qwen3-VL-4B, we branch from a common checkpoint after two supervised epochs and two epochs in the first self-play round. Both branches continue from identical model weights. The fixed branch retains the first round's opponent, generated readings and masks. The refreshed branch uses the common checkpoint as its new opponent, regenerates the readings and recomputes the masks and opponent scores.

\noindent Both branches use the same 60k training pages for two further epochs, with matched batch order, optimizer settings and update counts. We compare their endpoints after four cumulative self-play epochs and report the shared supervised stage separately. Computation includes generation, alignment and scoring, so equal update counts do not imply equal cost. This test concerns the joint refresh of the opponent and training records, not regenerated readings alone.

\noindent \textbf{Minimal controls.} We use two additional controls on Qwen3-VL-4B from the same SFT-2 checkpoint. Continued SFT uses three two-epoch blocks with full-reference negative log-likelihood. The masking control scores every response position in the contrastive term while retaining discrepancy-based FFL and its coefficient. It uses three two-epoch rounds and refreshes its own readings between rounds. We also compare first-round checkpoints, where the self-play variants share generated readings and FFL masks. Later rounds compare complete training procedures because their readings and masks may diverge.

\noindent For each of three training seeds, the branches share that seed's SFT-2 checkpoint. Data splits, stage endpoints and the optimizer reset policy are fixed across comparisons. We use the frozen Vary-600K and DocVQA evaluation sets and a fixed 5k-page DocBank test subset, keeping this subset separate from full-test results. We retain per-seed scores and measure actual update counts and cumulative computation.

\noindent \textbf{Architecture comparisons.} We compare language-model tuning without OCR, joint language-model and OCR tuning, and OCR-only modular tuning. The visual encoder and its connector remain frozen in all three variants. Each variant completes its own two-epoch SFT stage before self-play, using identical pretrained weights for shared modules. The OCR-equipped variants share the OCR initialization and fixed external OCR outputs. This comparison concerns complete adaptation strategies rather than changing trainable modules only after a shared SFT checkpoint.

\noindent \textbf{Training-cost measurement.} We record elapsed time and allocated GPU-hours from local input preparation and model loading to the saved checkpoint. The ledger includes external OCR, supervised training, each round's generation, alignment, opponent scoring, optimization and required I/O. We record stage and cumulative costs separately, count shared preprocessing once per standalone adaptation run, and retain a separate ledger of actual project-wide computation. Evaluation and startup costs are recorded separately.

\noindent \textbf{Cost comparison.} We record costs for OCR-only and joint language-model and OCR tuning on Qwen3-VL-4B. Both retain the frozen visual pathway. We compare their SFT-2 and SP-DR-3 checkpoints under the same data and input protocol. For normalized costs, we use the complete joint-tuning SFT-2 adaptation run as the denominator. Each quality score must use the metric protocol and checkpoint corresponding to its cost.

\noindent \textbf{Inference measurement.} We perform a batch-one comparison between the complete reader and its backbone without OCR on one H20 GPU. The same fixed English and Chinese pages use matched image and decoding settings. End-to-end latency includes page processing, external OCR, encoding and generation after model loading, with startup recorded separately. We record output lengths, response failures, latency distributions, throughput and peak memory. A separate cached-OCR measurement does not replace the online result.

\begin{samepage}
\noindent \textbf{Deterministic token alignment.} We compute LCS lengths for all suffix pairs before recovering matches. Let $n=|y|$ and $m=|\hat{y}|$, and let $D_{i,j}$ be the LCS length of the suffixes starting at positions $i$ and $j$. We set $D_{n+1,j}=D_{i,m+1}=0$ and compute
\begin{equation}\label{eq:lcs_suffix}
D_{i,j}=
\begin{cases}
1+D_{i+1,j+1}, & y_i=\hat{y}_j,\\
\max(D_{i+1,j},D_{i,j+1}), & y_i\ne\hat{y}_j.
\end{cases}
\end{equation}
\end{samepage}
Starting at $(i,j)=(1,1)$, we match equal tokens at their original positions and increment both indices. At unequal tokens, we increment $i$ only when $D_{i+1,j}>D_{i,j+1}$. Otherwise, we increment $j$, so ties skip the synthetic token while retaining the ground-truth position. Each step preserves an optimal LCS for the remaining suffixes. Recovery stops when either sequence ends. All unmatched original indices, including trailing positions, form $\mathcal{E}_y$ and $\mathcal{E}_{\hat{y}}$. Scoring retains each sequence's full original prefix. Matched EOS tokens are masked out. If $n<1{,}024$ and the synthetic reading lacks EOS, the ground-truth EOS contributes to RDM and FFL. We add no separate EOS loss.

\noindent \textbf{Empty masks and batch averaging.} An empty position set has score zero in \eqref{eq:masked_score}. Pairs with one empty mask are retained. FFL is zero whenever $\mathcal{E}_y$ is empty. With both masks empty, \eqref{eq:rdm} gives the constant $\log 2$ and \eqref{eq:ffl} gives zero, so neither term contributes a gradient. For a batch of $B$ original samples, let $r_b$ and $h_b$ be the per-sample RDM and FFL losses. We use
\begin{equation}\label{eq:batch_average}
\mathcal{L}_{\mathrm{batch}}=\frac{1}{B}\sum_{b=1}^{B}(r_b+\lambda h_b).
\end{equation}
The denominator remains $B$, including samples with both masks empty.

\begin{table*}[t]
\caption{Full reading accuracy on the Vary-600K validation set across all supervised epochs and self-play rounds. Each cell shows the change over the previous row, and SP-DR-1 is measured against SFT-2. Improvements are in red and regressions in blue. AVG is the mean of NED, Token-F1, ROUGE-L, and BLEU-4 on a 0 to 100 scale. SP-DR-3 is the model reported in the main text.}

\centering
\scriptsize
\resizebox{\textwidth}{!}{%
\begin{tabular}{clccccccc}
\toprule
Models & Method & CER(\%)$\downarrow$ & WER(\%)$\downarrow$ & NED$\uparrow$ & Token-F1$\uparrow$ & ROUGE-L$\uparrow$ & BLEU-4$\uparrow$ & AVG$\uparrow$ \\
\midrule
\multirow{9}{*}{\begin{tabular}[c]{@{}c@{}}InternVL\\3.5-4B\end{tabular}}
& Base & 4.85 & 9.12 & 0.938 & 0.941 & 87.29 & 80.44 & 88.91 \\
& SFT-1 & 2.31{\tiny (\textcolor{red}{-2.54})} & 4.60{\tiny (\textcolor{red}{-4.52})} & 0.969{\tiny (\textcolor{red}{+0.031})} & 0.972{\tiny (\textcolor{red}{+0.031})} & 92.87{\tiny (\textcolor{red}{+5.58})} & 88.32{\tiny (\textcolor{red}{+7.88})} & 93.82{\tiny (\textcolor{red}{+4.91})} \\
& SFT-2 & 2.12{\tiny (\textcolor{red}{-0.19})} & 4.00{\tiny (\textcolor{red}{-0.60})} & 0.971{\tiny (\textcolor{red}{+0.002})} & 0.975{\tiny (\textcolor{red}{+0.003})} & 93.39{\tiny (\textcolor{red}{+0.52})} & 89.04{\tiny (\textcolor{red}{+0.72})} & 94.26{\tiny (\textcolor{red}{+0.44})} \\
& SFT-3 & 2.34{\tiny (\textcolor{blue}{+0.22})} & 4.37{\tiny (\textcolor{blue}{+0.37})} & 0.968{\tiny (\textcolor{blue}{-0.003})} & 0.976{\tiny (\textcolor{red}{+0.001})} & 92.93{\tiny (\textcolor{blue}{-0.46})} & 88.60{\tiny (\textcolor{blue}{-0.44})} & 93.98{\tiny (\textcolor{blue}{-0.28})} \\
& SP-DR-1 & 1.77{\tiny (\textcolor{red}{-0.35})} & 3.49{\tiny (\textcolor{red}{-0.51})} & 0.978{\tiny (\textcolor{red}{+0.007})} & 0.982{\tiny (\textcolor{red}{+0.007})} & 95.28{\tiny (\textcolor{red}{+1.89})} & 91.43{\tiny (\textcolor{red}{+2.39})} & 95.68{\tiny (\textcolor{red}{+1.42})} \\
& SP-DR-2 & 1.35{\tiny (\textcolor{red}{-0.42})} & 2.83{\tiny (\textcolor{red}{-0.66})} & 0.984{\tiny (\textcolor{red}{+0.006})} & 0.989{\tiny (\textcolor{red}{+0.007})} & 96.77{\tiny (\textcolor{red}{+1.49})} & 93.01{\tiny (\textcolor{red}{+1.58})} & 96.77{\tiny (\textcolor{red}{+1.09})} \\
& SP-DR-3 & 1.17{\tiny (\textcolor{red}{-0.18})} & 2.35{\tiny (\textcolor{red}{-0.48})} & 0.988{\tiny (\textcolor{red}{+0.004})} & 0.991{\tiny (\textcolor{red}{+0.002})} & 97.90{\tiny (\textcolor{red}{+1.13})} & 94.52{\tiny (\textcolor{red}{+1.51})} & 97.58{\tiny (\textcolor{red}{+0.81})} \\
& SP-DR-4 & 1.13{\tiny (\textcolor{red}{-0.04})} & 2.24{\tiny (\textcolor{red}{-0.11})} & 0.989{\tiny (\textcolor{red}{+0.001})} & 0.992{\tiny (\textcolor{red}{+0.001})} & 98.06{\tiny (\textcolor{red}{+0.16})} & 94.79{\tiny (\textcolor{red}{+0.27})} & 97.74{\tiny (\textcolor{red}{+0.16})} \\
& SP-DR-5 & 1.11{\tiny (\textcolor{red}{-0.02})} & 2.20{\tiny (\textcolor{red}{-0.04})} & 0.990{\tiny (\textcolor{red}{+0.001})} & 0.993{\tiny (\textcolor{red}{+0.001})} & 98.13{\tiny (\textcolor{red}{+0.07})} & 94.88{\tiny (\textcolor{red}{+0.09})} & 97.83{\tiny (\textcolor{red}{+0.09})} \\
\midrule
\multirow{9}{*}{\begin{tabular}[c]{@{}c@{}}Qwen3\\VL-4B\end{tabular}}
& Base & 5.22 & 9.89 & 0.931 & 0.935 & 86.43 & 79.11 & 88.04 \\
& SFT-1 & 2.50{\tiny (\textcolor{red}{-2.72})} & 5.17{\tiny (\textcolor{red}{-4.72})} & 0.964{\tiny (\textcolor{red}{+0.033})} & 0.968{\tiny (\textcolor{red}{+0.033})} & 91.99{\tiny (\textcolor{red}{+5.56})} & 87.24{\tiny (\textcolor{red}{+8.13})} & 93.11{\tiny (\textcolor{red}{+5.07})} \\
& SFT-2 & 2.40{\tiny (\textcolor{red}{-0.10})} & 4.94{\tiny (\textcolor{red}{-0.23})} & 0.967{\tiny (\textcolor{red}{+0.003})} & 0.971{\tiny (\textcolor{red}{+0.003})} & 92.63{\tiny (\textcolor{red}{+0.64})} & 88.05{\tiny (\textcolor{red}{+0.81})} & 93.62{\tiny (\textcolor{red}{+0.51})} \\
& SFT-3 & 2.66{\tiny (\textcolor{blue}{+0.26})} & 5.33{\tiny (\textcolor{blue}{+0.39})} & 0.962{\tiny (\textcolor{blue}{-0.005})} & 0.967{\tiny (\textcolor{blue}{-0.004})} & 91.88{\tiny (\textcolor{blue}{-0.75})} & 87.43{\tiny (\textcolor{blue}{-0.62})} & 93.05{\tiny (\textcolor{blue}{-0.57})} \\
& SP-DR-1 & 1.82{\tiny (\textcolor{red}{-0.58})} & 4.06{\tiny (\textcolor{red}{-0.88})} & 0.9835{\tiny (\textcolor{red}{+0.0165})} & 0.978{\tiny (\textcolor{red}{+0.007})} & 94.47{\tiny (\textcolor{red}{+1.84})} & 90.50{\tiny (\textcolor{red}{+2.45})} & 95.28{\tiny (\textcolor{red}{+1.66})} \\
& SP-DR-2 & 1.45{\tiny (\textcolor{red}{-0.37})} & 3.30{\tiny (\textcolor{red}{-0.76})} & 0.988{\tiny (\textcolor{red}{+0.0045})} & 0.987{\tiny (\textcolor{red}{+0.009})} & 96.29{\tiny (\textcolor{red}{+1.82})} & 92.63{\tiny (\textcolor{red}{+2.13})} & 96.61{\tiny (\textcolor{red}{+1.33})} \\
& SP-DR-3 & 1.11{\tiny (\textcolor{red}{-0.34})} & 2.39{\tiny (\textcolor{red}{-0.91})} & 0.987{\tiny (\textcolor{blue}{-0.001})} & 0.992{\tiny (\textcolor{red}{+0.005})} & 97.74{\tiny (\textcolor{red}{+1.45})} & 94.49{\tiny (\textcolor{red}{+1.86})} & 97.53{\tiny (\textcolor{red}{+0.93})} \\
& SP-DR-4 & 1.09{\tiny (\textcolor{red}{-0.02})} & 2.15{\tiny (\textcolor{red}{-0.24})} & 0.989{\tiny (\textcolor{red}{+0.002})} & 0.994{\tiny (\textcolor{red}{+0.002})} & 98.02{\tiny (\textcolor{red}{+0.28})} & 94.78{\tiny (\textcolor{red}{+0.29})} & 97.78{\tiny (\textcolor{red}{+0.24})} \\
& SP-DR-5 & 1.06{\tiny (\textcolor{red}{-0.03})} & 2.18{\tiny (\textcolor{blue}{+0.03})} & 0.990{\tiny (\textcolor{red}{+0.001})} & 0.994{\tiny (0.000)} & 98.10{\tiny (\textcolor{red}{+0.08})} & 94.86{\tiny (\textcolor{red}{+0.08})} & 97.84{\tiny (\textcolor{red}{+0.06})} \\
\bottomrule
\end{tabular}
}
\label{tab:vary_full}
\end{table*}

\begin{table*}[t]
\caption{Full zero-shot trajectories on DocBank, IIT-CDIP, DocVQA, and InfographicVQA. Each cell shows the change over the previous row, and SP-DR-1 is measured against SFT-2. Improvements are in red and regressions in blue. AVG is the mean of the two NED scores and the two ANLS scores on a 0 to 100 scale. SP-DR-3 is the model reported in the main text.}
\centering
\scriptsize
\resizebox{\textwidth}{!}{%
\begin{tabular}{clccccccc}
\toprule
\multirow{2}{*}{Models} & \multirow{2}{*}{Method} & \multicolumn{2}{c}{DocBank} & \multicolumn{2}{c}{IIT-CDIP} & \multicolumn{2}{c}{VQA (ANLS$\uparrow$)} & \multirow{2}{*}{AVG$\uparrow$} \\
\cmidrule(lr){3-4}\cmidrule(lr){5-6}\cmidrule(lr){7-8}
& & CER(\%)$\downarrow$ & NED$\uparrow$ & CER(\%)$\downarrow$ & NED$\uparrow$ & DocVQA & InfoVQA & \\
\midrule
\multirow{9}{*}{\begin{tabular}[c]{@{}c@{}}InternVL\\3.5-4B\end{tabular}}
& Base & 13.44 & 0.812 & 19.56 & 0.735 & 72.79 & 50.52 & 69.50 \\
& SFT-1 & 12.40{\tiny (\textcolor{red}{-1.04})} & 0.835{\tiny (\textcolor{red}{+0.023})} & 18.39{\tiny (\textcolor{red}{-1.17})} & 0.762{\tiny (\textcolor{red}{+0.027})} & 72.71{\tiny (\textcolor{blue}{-0.08})} & 49.20{\tiny (\textcolor{blue}{-1.32})} & 70.40{\tiny (\textcolor{red}{+0.90})} \\
& SFT-2 & 11.60{\tiny (\textcolor{red}{-0.80})} & 0.848{\tiny (\textcolor{red}{+0.013})} & 17.83{\tiny (\textcolor{red}{-0.56})} & 0.778{\tiny (\textcolor{red}{+0.016})} & 73.01{\tiny (\textcolor{red}{+0.30})} & 48.15{\tiny (\textcolor{blue}{-1.05})} & 70.94{\tiny (\textcolor{red}{+0.54})} \\
& SFT-3 & 12.17{\tiny (\textcolor{blue}{+0.57})} & 0.829{\tiny (\textcolor{blue}{-0.019})} & 18.44{\tiny (\textcolor{blue}{+0.61})} & 0.762{\tiny (\textcolor{blue}{-0.016})} & 72.52{\tiny (\textcolor{blue}{-0.49})} & 48.24{\tiny (\textcolor{red}{+0.09})} & 69.97{\tiny (\textcolor{blue}{-0.97})} \\
& SP-DR-1 & 10.55{\tiny (\textcolor{red}{-1.05})} & 0.866{\tiny (\textcolor{red}{+0.018})} & 16.70{\tiny (\textcolor{red}{-1.13})} & 0.802{\tiny (\textcolor{red}{+0.024})} & 73.48{\tiny (\textcolor{red}{+0.47})} & 48.40{\tiny (\textcolor{red}{+0.25})} & 72.17{\tiny (\textcolor{red}{+1.23})} \\
& SP-DR-2 & 10.08{\tiny (\textcolor{red}{-0.47})} & 0.876{\tiny (\textcolor{red}{+0.010})} & 15.96{\tiny (\textcolor{red}{-0.74})} & 0.813{\tiny (\textcolor{red}{+0.011})} & 73.92{\tiny (\textcolor{red}{+0.44})} & 48.22{\tiny (\textcolor{blue}{-0.18})} & 72.76{\tiny (\textcolor{red}{+0.59})} \\
& SP-DR-3 & 9.73{\tiny (\textcolor{red}{-0.35})} & 0.884{\tiny (\textcolor{red}{+0.008})} & 15.35{\tiny (\textcolor{red}{-0.61})} & 0.823{\tiny (\textcolor{red}{+0.010})} & 74.27{\tiny (\textcolor{red}{+0.35})} & 48.06{\tiny (\textcolor{blue}{-0.16})} & 73.26{\tiny (\textcolor{red}{+0.50})} \\
& SP-DR-4 & 9.58{\tiny (\textcolor{red}{-0.15})} & 0.888{\tiny (\textcolor{red}{+0.004})} & 15.08{\tiny (\textcolor{red}{-0.27})} & 0.827{\tiny (\textcolor{red}{+0.004})} & 74.41{\tiny (\textcolor{red}{+0.14})} & 48.02{\tiny (\textcolor{blue}{-0.04})} & 73.48{\tiny (\textcolor{red}{+0.23})} \\
& SP-DR-5 & 9.55{\tiny (\textcolor{red}{-0.03})} & 0.887{\tiny (\textcolor{blue}{-0.001})} & 15.06{\tiny (\textcolor{red}{-0.02})} & 0.826{\tiny (\textcolor{blue}{-0.001})} & 74.39{\tiny (\textcolor{blue}{-0.02})} & 47.98{\tiny (\textcolor{blue}{-0.04})} & 73.42{\tiny (\textcolor{blue}{-0.06})} \\
\midrule
\multirow{9}{*}{\begin{tabular}[c]{@{}c@{}}Qwen3\\VL-4B\end{tabular}}
& Base & 12.82 & 0.834 & 18.45 & 0.756 & 78.54 & 52.21 & 72.44 \\
& SFT-1 & 11.58{\tiny (\textcolor{red}{-1.24})} & 0.857{\tiny (\textcolor{red}{+0.023})} & 17.34{\tiny (\textcolor{red}{-1.11})} & 0.788{\tiny (\textcolor{red}{+0.032})} & 80.05{\tiny (\textcolor{red}{+1.51})} & 54.12{\tiny (\textcolor{red}{+1.91})} & 74.67{\tiny (\textcolor{red}{+2.23})} \\
& SFT-2 & 10.95{\tiny (\textcolor{red}{-0.63})} & 0.868{\tiny (\textcolor{red}{+0.011})} & 16.71{\tiny (\textcolor{red}{-0.63})} & 0.802{\tiny (\textcolor{red}{+0.014})} & 81.22{\tiny (\textcolor{red}{+1.17})} & 55.34{\tiny (\textcolor{red}{+1.22})} & 75.89{\tiny (\textcolor{red}{+1.22})} \\
& SFT-3 & 11.42{\tiny (\textcolor{blue}{+0.47})} & 0.852{\tiny (\textcolor{blue}{-0.016})} & 16.25{\tiny (\textcolor{red}{-0.46})} & 0.803{\tiny (\textcolor{red}{+0.001})} & 80.45{\tiny (\textcolor{blue}{-0.77})} & 54.52{\tiny (\textcolor{blue}{-0.82})} & 75.12{\tiny (\textcolor{blue}{-0.77})} \\
& SP-DR-1 & 10.02{\tiny (\textcolor{red}{-0.93})} & 0.8798{\tiny (\textcolor{red}{+0.0118})} & 15.92{\tiny (\textcolor{red}{-0.79})} & 0.814{\tiny (\textcolor{red}{+0.012})} & 83.08{\tiny (\textcolor{red}{+1.86})} & 56.42{\tiny (\textcolor{red}{+1.08})} & 77.22{\tiny (\textcolor{red}{+1.33})} \\
& SP-DR-2 & 9.44{\tiny (\textcolor{red}{-0.58})} & 0.886{\tiny (\textcolor{red}{+0.0062})} & 15.04{\tiny (\textcolor{red}{-0.88})} & 0.829{\tiny (\textcolor{red}{+0.015})} & 83.61{\tiny (\textcolor{red}{+0.53})} & 57.65{\tiny (\textcolor{red}{+1.23})} & 78.19{\tiny (\textcolor{red}{+0.97})} \\
& SP-DR-3 & 9.05{\tiny (\textcolor{red}{-0.39})} & 0.898{\tiny (\textcolor{red}{+0.012})} & 14.42{\tiny (\textcolor{red}{-0.62})} & 0.847{\tiny (\textcolor{red}{+0.018})} & 84.89{\tiny (\textcolor{red}{+1.28})} & 58.94{\tiny (\textcolor{red}{+1.29})} & 79.58{\tiny (\textcolor{red}{+1.39})} \\
& SP-DR-4 & 8.95{\tiny (\textcolor{red}{-0.10})} & 0.904{\tiny (\textcolor{red}{+0.006})} & 14.01{\tiny (\textcolor{red}{-0.41})} & 0.845{\tiny (\textcolor{blue}{-0.002})} & 84.83{\tiny (\textcolor{blue}{-0.06})} & 59.51{\tiny (\textcolor{red}{+0.57})} & 79.81{\tiny (\textcolor{red}{+0.23})} \\
& SP-DR-5 & 8.98{\tiny (\textcolor{blue}{+0.03})} & 0.900{\tiny (\textcolor{blue}{-0.004})} & 14.00{\tiny (\textcolor{red}{-0.01})} & 0.845{\tiny (0.000)} & 84.85{\tiny (\textcolor{red}{+0.02})} & 59.88{\tiny (\textcolor{red}{+0.37})} & 79.81{\tiny (0.00)} \\
\bottomrule
\end{tabular}
}
\label{tab:zeroshot_full}
\end{table*}

\section{Benchmarks and Data Sources}
\label{app:benchmarks}

\begin{itemize}
\item \textbf{Vary-600K} is a released subset of the pretraining data used in Vary \cite{vary_dataset}.\footnote{\url{https://github.com/Ucas-HaoranWei/Vary-tiny-600k}} It contains approximately 600{,}000 PDF page images paired with extracted text. We use it for the supervised stage and for in-distribution evaluation.
\item \textbf{DocBank} \cite{docbank} contains 500{,}000 pages from arXiv, where each token carries a semantic label and a box. We use its test set for zero-shot academic recognition. Full-page transcription evaluation also requires a defined reading order and text serialization.
\item \textbf{IIT-CDIP} \cite{iit_cdip} contains millions of scanned documents made public through tobacco-industry litigation. We use it to test robustness to scanning noise, scored by CER and NED. Reference verification is described below.
\item \textbf{DocVQA} \cite{mathew2021docvqa} poses questions about individual document images. Answering can require reading text and interpreting its layout. We report ANLS.
\item \textbf{InfographicVQA} \cite{mathew2022infographicvqa} targets infographics that mix dense text with charts and icons. It probes reading under varied and non-linear layouts.
\item \textbf{OmniDocBench} \cite{ouyang2024omnidocbench} evaluates parsing across diverse PDFs with rich annotations. Table~\ref{tab:ocrcompare} reports Text Edit Distance.
\item \textbf{OCRBench v2} \cite{fu2024ocrbenchv2} measures how well a model locates text and reasons over it across many subtasks. It reports separate results for its English and Chinese subsets.
\item \textbf{CC-OCR} \cite{yang2024ccocr} spans multi-scene reading, multilingual reading, parsing, and key information extraction. It uses task-specific metrics. Our cross-track summary is described below.
\end{itemize}

\noindent \textbf{Data audit.} We fix the dataset revisions, split manifests, reference files and evaluation settings. We record both question and image counts for VQA tasks and distinguish public corpus sizes from the examples actually evaluated. We keep target-benchmark data out of training and model selection and use no task demonstrations at inference. Development and final evaluation pages are separated from training at the source-document level wherever provenance permits.

\noindent \textbf{Reference verification.} We verify the provenance and accuracy of the IIT-CDIP evaluation transcriptions before using them as references for CER and NED.

\noindent \textbf{Benchmark scoring.} For DocVQA and InfographicVQA, we compute ANLS by taking the best thresholded edit similarity over each question's allowed answers, then averaging over questions on a 0 to 100 scale. The scorer lowercases answers and requires normalized edit distance below 0.5. For OmniDocBench, OCRBench v2 and CC-OCR, we fix the dataset and evaluator revisions and retain their task-specific processing. We use Text Edit Distance for OmniDocBench, an equal-language mean for OCRBench v2, and an equal-track mean for CC-OCR, with the underlying language and track scores retained.

\section{Full Iteration Trajectories}
\label{app:trajectories}

\noindent Tables~\ref{tab:vary_full} and~\ref{tab:zeroshot_full} give the supervised and self-play trajectories. Improvements are not monotonic across all metrics. For example, Qwen3-VL-4B's Vary-600K NED changes from 0.988 at SP-DR-2 to 0.987 at SP-DR-3. The main comparison uses the third-round endpoint rather than selecting a different round for each metric.

\noindent \textbf{Aggregate score scope.} The averages summarize different sets of metrics. Table~\ref{tab:vary_full} combines NED, Token-F1, ROUGE-L and BLEU-4. Tables~\ref{tab:zeroshot_full} and~\ref{tab:modality} combine the two zero-shot NED scores and two ANLS scores. Figure~\ref{fig:analysis}(c) uses this four-task mean for the base, SFT-2 and third-round checkpoints. Figure~\ref{fig:epochs} combines Vary-600K NED, DocBank NED and DocVQA ANLS. Each mean uses a 0 to 100 scale. Figure~\ref{fig:analysis}(a) uses the equal-weight mean of Vary-600K, DocBank and IIT-CDIP NED, each multiplied by 100, and DocVQA and InfographicVQA ANLS. This five-task mean includes in-distribution performance on Vary-600K, whereas panel (c) includes only the four zero-shot tasks. The two averages therefore measure different evaluation scopes and should not be compared directly.

\noindent \textbf{Reported precision.} AVG values and their changes are displayed to two decimal places. A change may differ by 0.01 from subtracting the displayed averages because of rounding.

\section{Modality-Specific Analysis}
\label{app:modality}

\noindent Table~\ref{tab:modality} compares full-tuning and modular trajectories. Across the reported iterations, modular tuning improves the four-task zero-shot AVG from 75.89 to 79.58, while full tuning changes from 75.92 to 74.98. The two trajectories show the advantage of the modular configuration on these evaluated tasks. The aggregate scores use the metric set specified in Appendix~\ref{app:trajectories}.

\section{Case Study}
\label{app:cases}

\noindent The selected Vary-600K cases compare supervised readings with third-round InternVL-3.5-4B outputs. Tables~\ref{tab:case-quicksort} and~\ref{tab:case-summa} show the supplied references and corresponding model readings. Red shading marks differences from the displayed references, including punctuation and spelling. These examples show corrected words and remaining errors in the displayed readings.

\begin{table*}[t]
\caption{Modality-specific analysis on Qwen3-VL-4B. Full tuning updates the backbone and the OCR module, while the modular design freezes the backbone. The Vary-600K column measures in-distribution reading, and AVG averages the four zero-shot scores on a 0 to 100 scale. Each cell shows the change over the previous row, with improvements in red and regressions in blue.}
\centering
\scriptsize
\resizebox{\textwidth}{!}{%
\begin{tabular}{clcccccc}
\toprule
Config & Method & Vary CER(\%)$\downarrow$ & DocBank NED$\uparrow$ & IIT-CDIP NED$\uparrow$ & DocVQA$\uparrow$ & InfoVQA$\uparrow$ & AVG$\uparrow$ \\
\midrule
\multirow{6}{*}{\begin{tabular}[c]{@{}c@{}}Full\\tuning\end{tabular}}
& SFT-2 & 2.25 & 0.872 & 0.805 & 81.00 & 55.00 & 75.92 \\
& Iter-1 & 1.55{\tiny (\textcolor{red}{-0.70})} & 0.878{\tiny (\textcolor{red}{+0.006})} & 0.812{\tiny (\textcolor{red}{+0.007})} & 81.50{\tiny (\textcolor{red}{+0.50})} & 55.30{\tiny (\textcolor{red}{+0.30})} & 76.45{\tiny (\textcolor{red}{+0.53})} \\
& Iter-2 & 1.18{\tiny (\textcolor{red}{-0.37})} & 0.871{\tiny (\textcolor{blue}{-0.007})} & 0.804{\tiny (\textcolor{blue}{-0.008})} & 81.10{\tiny (\textcolor{blue}{-0.40})} & 54.60{\tiny (\textcolor{blue}{-0.70})} & 75.80{\tiny (\textcolor{blue}{-0.65})} \\
& Iter-3 & 0.98{\tiny (\textcolor{red}{-0.20})} & 0.862{\tiny (\textcolor{blue}{-0.009})} & 0.795{\tiny (\textcolor{blue}{-0.009})} & 80.41{\tiny (\textcolor{blue}{-0.69})} & 53.80{\tiny (\textcolor{blue}{-0.80})} & 74.98{\tiny (\textcolor{blue}{-0.82})} \\
& Iter-4 & 0.95{\tiny (\textcolor{red}{-0.03})} & 0.851{\tiny (\textcolor{blue}{-0.011})} & 0.786{\tiny (\textcolor{blue}{-0.009})} & 79.80{\tiny (\textcolor{blue}{-0.61})} & 53.10{\tiny (\textcolor{blue}{-0.70})} & 74.15{\tiny (\textcolor{blue}{-0.83})} \\
& Iter-5 & 0.94{\tiny (\textcolor{red}{-0.01})} & 0.843{\tiny (\textcolor{blue}{-0.008})} & 0.779{\tiny (\textcolor{blue}{-0.007})} & 79.50{\tiny (\textcolor{blue}{-0.30})} & 52.70{\tiny (\textcolor{blue}{-0.40})} & 73.60{\tiny (\textcolor{blue}{-0.55})} \\
\midrule
\multirow{6}{*}{\begin{tabular}[c]{@{}c@{}}Modular\\(ours)\end{tabular}}
& SFT-2 & 2.40 & 0.868 & 0.802 & 81.22 & 55.34 & 75.89 \\
& Iter-1 & 1.82{\tiny (\textcolor{red}{-0.58})} & 0.8798{\tiny (\textcolor{red}{+0.0118})} & 0.814{\tiny (\textcolor{red}{+0.012})} & 83.08{\tiny (\textcolor{red}{+1.86})} & 56.42{\tiny (\textcolor{red}{+1.08})} & 77.22{\tiny (\textcolor{red}{+1.33})} \\
& Iter-2 & 1.45{\tiny (\textcolor{red}{-0.37})} & 0.886{\tiny (\textcolor{red}{+0.0062})} & 0.829{\tiny (\textcolor{red}{+0.015})} & 83.61{\tiny (\textcolor{red}{+0.53})} & 57.65{\tiny (\textcolor{red}{+1.23})} & 78.19{\tiny (\textcolor{red}{+0.97})} \\
& Iter-3 & 1.11{\tiny (\textcolor{red}{-0.34})} & 0.898{\tiny (\textcolor{red}{+0.012})} & 0.847{\tiny (\textcolor{red}{+0.018})} & 84.89{\tiny (\textcolor{red}{+1.28})} & 58.94{\tiny (\textcolor{red}{+1.29})} & 79.58{\tiny (\textcolor{red}{+1.39})} \\
& Iter-4 & 1.09{\tiny (\textcolor{red}{-0.02})} & 0.904{\tiny (\textcolor{red}{+0.006})} & 0.845{\tiny (\textcolor{blue}{-0.002})} & 84.83{\tiny (\textcolor{blue}{-0.06})} & 59.51{\tiny (\textcolor{red}{+0.57})} & 79.81{\tiny (\textcolor{red}{+0.23})} \\
& Iter-5 & 1.06{\tiny (\textcolor{red}{-0.03})} & 0.900{\tiny (\textcolor{blue}{-0.004})} & 0.845{\tiny (0.000)} & 84.85{\tiny (\textcolor{red}{+0.02})} & 59.88{\tiny (\textcolor{red}{+0.37})} & 79.81{\tiny (0.00)} \\
\bottomrule
\end{tabular}
}
\label{tab:modality}
\end{table*}

\begin{table*}[p]
    \caption{Vary-600K Example 1, a Quicksort page, with the supplied reference text and reported readings. The reference does not cover the full page shown in the thumbnail. Red shading marks differences from the displayed reference without changing the readings.}
    \label{tab:case-quicksort}
    \centering
    \resizebox{\textwidth}{!}{%
    \begin{tabular}{@{}|>{\raggedright\arraybackslash}p{0.18\textwidth}|>{\raggedright\arraybackslash}p{0.80\textwidth}|@{}}
        \hline
        \parbox[t]{\linewidth}{\vspace{0pt}\includegraphics[width=1.0\linewidth]{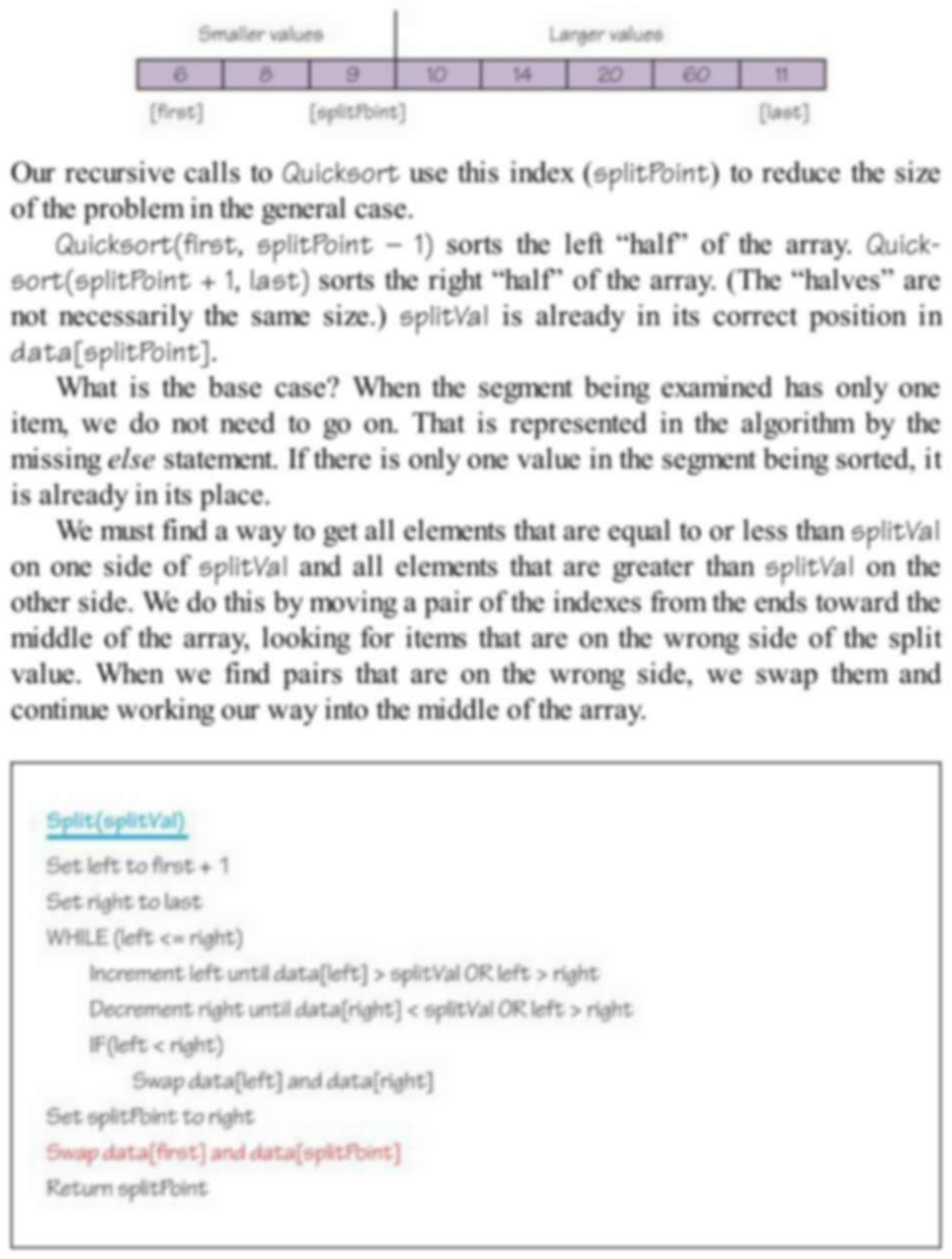}}
        Vary 600K Example 1
        & 
        \parbox[t]{\linewidth}{\raggedright \scriptsize Ground Truth:"<lb><lb>Our recursive calls to Quicksort use this index (splitPoint) to reduce the size of the problem in the general case.<lb>Quicksort(first, splitPoint – 1) sorts the left “half” of the array. Quick-sort(splitPoint + 1, last) sorts the right “half” of the array. (The “halves” are not necessarily the same size.) splitVal is already in its correct position in data[splitPoint].<lb>What is the base case? When the segment being examined has only one item, we do not need to go on. That is represented in the algorithm by the missing else statement. If there is only one value in the segment being sorted, it is already in its place.<lb>We must find a way to get all elements that are equal to or less than splitVal on one side of splitVal and all elements that are greater than splitVal on the other side. We do this by moving a pair of the indexes from the ends toward the middle of the array, looking for items that are on the wrong side of the split value. When we find pairs that are on the wrong side, we swap them and continue working our way into the middle of the array.<lb>
        \texttt{\textbackslash begin\{tabular\}\{|c|\}} <lb>
        \texttt{\textbackslash hline\ \textbackslash\textbackslash} <lb>
        \texttt{\textbackslash end\{tabular\}}<lb>"} \\
        \hline
        \parbox[t]{\linewidth}{SFT-2}  
        & 
        \parbox[t]{\linewidth}{\raggedright \scriptsize Document Reading:"<lb><lb>Our \colorbox{red!20}{recurslve} calls to Quicksort use this index (splitPoint) to reduce the size of the problem in the \colorbox{red!20}{genral} case.<lb>
        Quicksort(first, splitPoint \colorbox{red!20}{-} 1) sorts the left “half” of the array. \colorbox{red!20}{Quicksort}(splitPoint + 1, last) sorts the right “half” of the \colorbox{red!20}{arry}. (The “halves” are not necessarily the same size.) \colorbox{red!20}{splitval} is already in its correct position in data[splitPoint].<lb>
        What is the base case? When the segment being \colorbox{red!20}{examned} has only one item, we do not need to go on. That is represented in the algorithm by the missing else statement. If there is only one value in the segment being sorted, it is already in its place.<lb>
        We must find a way to get all elements that are equal to or less than splitVal on one side of splitVal and all elements that are greater than splitVal on the other side. \colorbox{red!20}{we} do this by moving a pair of the \colorbox{red!20}{indices} from the ends toward the middle of the array, looking for items that are on the wrong side of the split value. When we find pairs that are on the wrong side, we swap them and continue working our way into the middle of the array.<lb>
        \texttt{\textbackslash begin\{tabular\}\{\colorbox{red!20}{|c}\}} <lb>
        \texttt{\textbackslash hline\ \textbackslash\textbackslash} <lb>
        \texttt{\textbackslash end\{tabular\}} <lb>
        "}
        \\
        \hline
        \parbox[t]{\linewidth}{SP-DocReader iteration 3}
        & 
        \parbox[t]{\linewidth}{\raggedright \scriptsize Document Reading:"<lb><lb>Our recursive calls to Quicksort use this index (splitPoint) to reduce the size of the problem in the general case.<lb>
        Quicksort(first, splitPoint \colorbox{red!20}{-} 1) sorts the left “half” of the array. \colorbox{red!20}{Quicksort}(splitPoint + 1, last) sorts the right “half” of the array. (The “halves” are not \colorbox{red!20}{necessarly} the same size.) splitVal is already in its correct position in data[splitPoint].<lb>
        What is the base case? When the segment being examined has only one item, we do not need to go on. That is represented in the algorithm by the missing else statement. If there is only one value in the segment being sorted, it is already in its place.<lb>
        We must find a way to get all elements that are equal to or less than splitVal on one side of splitVal and all elements that are greater than splitVal on the other side. We do this by moving a pair of the \colorbox{red!20}{indixes} from the ends toward the middle of the array, looking for items that are on the wrong side of the split value. When we find pairs that are on the wrong side, we swap them and continue working our way into the middle of the array.<lb>
        \texttt{\textbackslash begin\{tabular\}\{|c|\}} <lb>
        \texttt{\textbackslash hline\ \textbackslash\textbackslash} <lb>
        \texttt{\textbackslash end\{tabular\}}<lb>
        "} \\
        \hline
    \end{tabular}
    }
\end{table*}

\clearpage

\begingroup
\setlength{\tabcolsep}{4pt}
\setlength{\LTcapwidth}{\textwidth}
\begin{longtable}{@{}|>{\raggedright\arraybackslash}p{0.18\textwidth}|>{\raggedright\arraybackslash}p{0.78\textwidth}|@{}}
    \caption{Vary-600K Example 2, a Summa Theologica page, with the supplied reference text and reported readings. The thumbnail identifies the source page. Red shading marks differences from the displayed reference without changing the readings.}
    \label{tab:case-summa}\\
    \endfirsthead
    \multicolumn{2}{l}{\footnotesize Table~\ref{tab:case-summa} (continued)}\\
    \hline
    \endhead
    \hline
    \endfoot
    \endlastfoot
        \hline
        \parbox[t]{\linewidth}{\vspace{0pt}\includegraphics[width=1.0\linewidth]{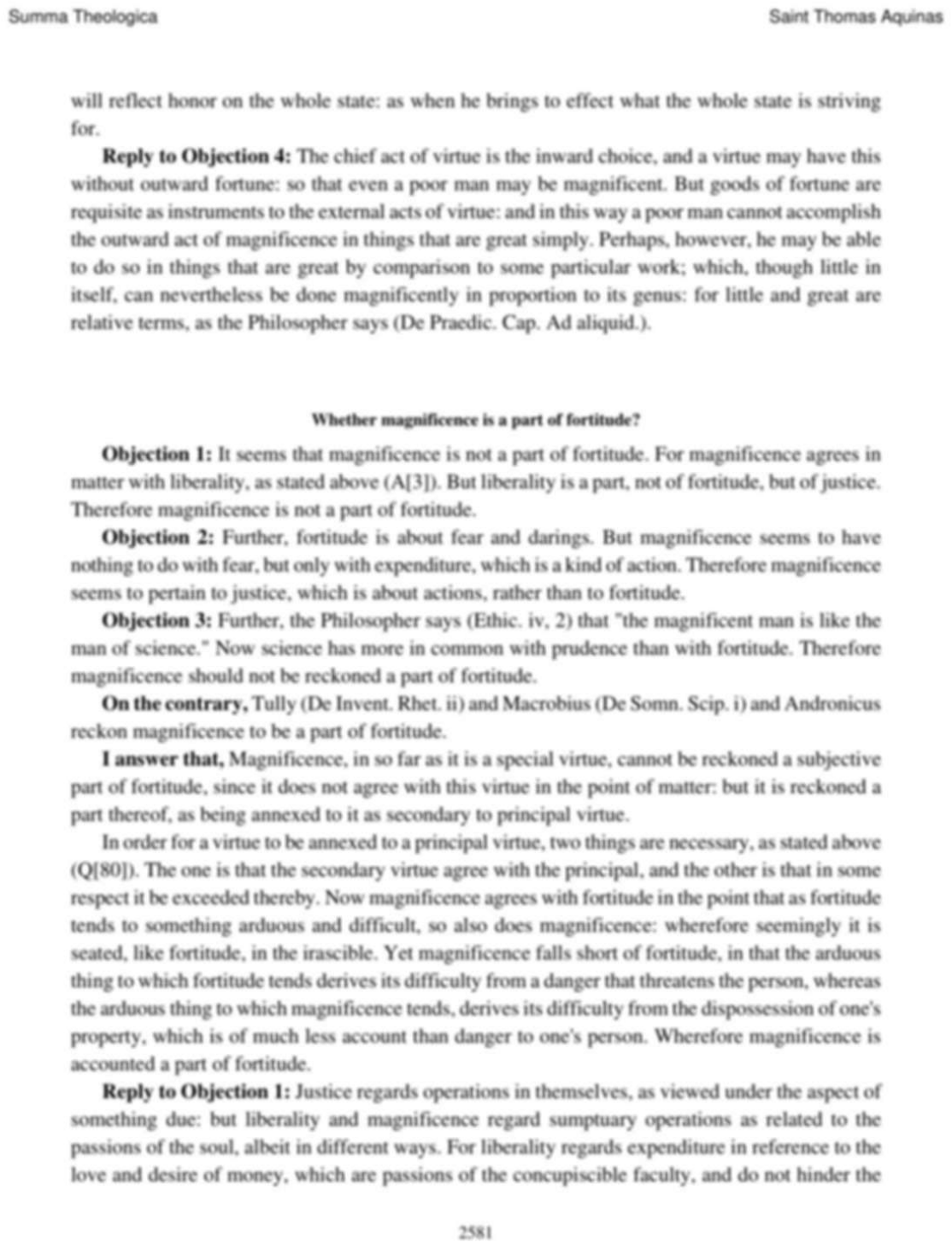}}
        Vary 600K Example 2
        & 
        \parbox[t]{\linewidth}{\raggedright \scriptsize Ground Truth:"<lb>Summa Theologica Saint Thomas Aquinas<lb>will reflect honor on the whole state: as when he brings to effect what the whole state is striving for.<lb>Reply to Objection 4: The chief act of virtue is the inward choice, and a virtue may have this without outward fortune: so that even a poor man may be magnificent. But goods of fortune are requisite as instruments to the external acts of virtue: and in this way a poor man cannot accomplish the outward act of magnificence in things that are great simply. Perhaps, however, he may be able to do so in things that are great by comparison to some particular work; which, though little in itself, can nevertheless be done magnificently in proportion to its genus: for little and great are relative terms, as the Philosopher says (De Praedic. Cap. Ad aliquid.).<lb>Whether magnificence is a part of fortitude?<lb>Objection 1: It seems that magnificence is not a part of fortitude. For magnificence agrees in matter with liberality, as stated above (A[3]). But liberality is a part, not of fortitude, but of justice. Therefore magnificence is not a part of fortitude.<lb>Objection 2: Further, fortitude is about fear and darings. But magnificence seems to have nothing to do with fear, but only with expenditure, which is a kind of action. Therefore magnificence seems to pertain to justice, which is about actions, rather than to fortitude.<lb>Objection 3: Further, the Philosopher says (Ethic. iv, 2) that "the magnificent man is like the man of science." Now science has more in common with prudence than with fortitude. Therefore magnificence should not be reckoned a part of fortitude.<lb>On the contrary, Tully (De Invent. Rhet. ii) and Macrobius (De Somn. Scip. i) and Andronicus reckon magnificence to be a part of fortitude.<lb>I answer that, Magnificence, in so far as it is a special virtue, cannot be reckoned a subjective part of fortitude, since it does not agree with this virtue in the point of matter: but it is reckoned a part thereof, as being annexed to it as secondary to principal virtue.<lb>In order for a virtue to be annexed to a principal virtue, two things are necessary, as stated above (Q[80]). The one is that the secondary virtue agree with the principal, and the other is that in some respect it be exceeded thereby. Now magnificence agrees with fortitude in the point that as fortitude tends to something arduous and difficult, so also does magnificence: wherefore seemingly it is seated, like fortitude, in the irascible. Yet magnificence falls short of fortitude, in that the arduous thing to which fortitude tends derives its difficulty from a danger that threatens the person, whereas the arduous thing to which magnificence tends, derives its difficulty from the dispossession of one's property, which is of much less account than danger to one's person. Wherefore magnificence is accounted a part of fortitude.<lb>Reply to Objection 1: Justice regards operations in themselves, as viewed under the aspect of something due: but liberality and magnificence regard sumptuary operations as related to the passions of the soul, albeit in different ways. For liberality regards expenditure in reference to the love and desire of money, which are passions of the concupiscible faculty, and do not hinder the<lb>2581<lb>"} \\
        \hline
        SFT-2 
        & 
        \parbox[t]{\linewidth}{\raggedright \scriptsize Document Reading:"<lb>Summa Theologica Saint Thomas Aquinas<lb>
will reflect honor on the whole state: as when he brings to \colorbox{red!20}{affect} what the whole state is striving for.<lb>
Reply to Objection 4: The chief act of virtue is the inward choice, and a virtue may have this without outward fortune: so that even a poor man may be magnificent. But goods of fortune are requisite as instruments to the external acts of virtue: and in this way a poor man cannot accomplish the outward act of magnificence in things that are great simply. Perhaps, however, he may be able to do so in things that are great by comparison to some particular work; which, though little in itself, can nevertheless be done magnificently in proportion to its genus: for little and great are relative terms, as the Philosopher says (De \colorbox{red!20}{Predic}. Cap. Ad \colorbox{red!20}{aliqid}.).<lb>
Whether magnificence is a part of fortitude?<lb>
Objection 1: It seems that magnificence is not a part of fortitude. For magnificence agrees in matter with liberality, as stated above (A[3]). But liberality is a part, not of fortitude, but of justice. Therefore magnificence is not a part of fortitude.<lb>
Objection 2: Further, fortitude is about fear and \colorbox{red!20}{daring}. But magnificence seems to have nothing to do with fear, but only with expenditure, which is a kind of action. Therefore magnificence seems to pertain to justice, which is about actions, rather than to fortitude.<lb>
Objection 3: Further, the Philosopher says (Ethic. iv, 2) that "the magnificent man is like the man of science." Now science has more in common with prudence than with fortitude. Therefore magnificence should not be reckoned a part of fortitude.<lb>
On the contrary, Tully (De Invent. Rhet. ii) and \colorbox{red!20}{Marcrobius} (De Somn. \colorbox{red!20}{Scp}. i) and Andronicus reckon magnificence to be a part of fortitude.<lb>
I answer that, Magnificence, in so far as it is a special virtue, cannot be \colorbox{red!20}{reckond} a subjective part of fortitude, since it does not agree with this virtue in the point of matter: but it is reckoned a part thereof, as being \colorbox{red!20}{anexed} to it as secondary to principal virtue.<lb>
In order for a virtue to be annexed to a principal virtue, two things are necessary, as stated above (Q[80]). The one is that the secondary virtue agree with the principal, and the other is that in some respect it be exceeded thereby. Now magnificence agrees with fortitude in the point that as fortitude tends to something arduous and difficult, so also does magnificence: wherefore seemingly it is seated, like fortitude, in the \colorbox{red!20}{irrascible}. Yet magnificence falls short of fortitude, in that the arduous thing to which fortitude tends derives its difficulty from a danger that threatens the person, whereas the arduous thing to which magnificence tends, derives its difficulty from the \colorbox{red!20}{dispossesion} of one's property, which is of much less \colorbox{red!20}{acount} than danger to one's person. Wherefore magnificence is accounted a part of fortitude.<lb>
Reply to Objection 1: Justice regards operations in themselves, as viewed under the aspect of something due: but liberality and magnificence regard \colorbox{red!20}{sumptary} operations as related to the passions of the soul, \colorbox{red!20}{albit} in different ways. For liberality regards expenditure in reference to the love and desire of money, which are passions of the \colorbox{red!20}{concupiscble} faculty, and do not hinder the<lb>
\colorbox{red!20}{258I}<lb>"}
        \\
        \hline
        \parbox[t]{\linewidth}{SP-DocReader iteration 3} 
        & 
        \parbox[t]{\linewidth}{\raggedright \scriptsize Document Reading:"<lb>Summa Theologica Saint Thomas Aquinas<lb>
will reflect honor on the whole state: as when he brings to effect what the whole state is striving for.<lb>
Reply to Objection 4: The chief act of virtue is the inward choice, and a virtue may have this without outward fortune: so that even a poor man may be magnificent. But goods of fortune are requisite as instruments to the external acts of virtue: and in this way a poor man cannot accomplish the outward act of magnificence in things that are great simply. Perhaps, however, he may be able to do so in things that are great by comparison to some particular work; which, though little in itself, can nevertheless be done magnificently in proportion to its genus: for little and great are relative terms, as the Philosopher says (De \colorbox{red!20}{Praedlc}. Cap. Ad \colorbox{red!20}{aliquld}.).<lb>
Whether magnificence is a part of fortitude?<lb>
Objection 1: It seems that magnificence is not a part of fortitude. For magnificence agrees in matter with liberality, as stated above (A[3]). But liberality is a part, not of fortitude, but of justice. Therefore magnificence is not a part of fortitude.<lb>
Objection 2: Further, fortitude is about fear and \colorbox{red!20}{daring}. But magnificence seems to have nothing to do with fear, but only with expenditure, which is a kind of action. Therefore magnificence seems to pertain to justice, which is about actions, rather than to fortitude.<lb>
Objection 3: Further, the Philosopher says (Ethic. iv, 2) that "the magnificent man is like the man of science." Now science has more in common with prudence than with fortitude. Therefore magnificence should not be reckoned a part of fortitude.<lb>
On the contrary, Tully (De Invent. Rhet. ii) and Macrobius (De Somn. \colorbox{red!20}{Scip. 1}) and Andronicus reckon magnificence to be a part of fortitude.<lb>
I answer that, Magnificence, in so far as it is a special virtue, cannot be reckoned a subjective part of fortitude, since it does not agree with this virtue in the point of matter: but it is reckoned a part thereof, as being annexed to it as secondary to principal virtue.<lb>
In order for a virtue to be annexed to a principal virtue, two things are necessary, as stated above (Q[80]). The one is that the secondary virtue agree with the principal, and the other is that in some respect it be exceeded thereby. Now magnificence agrees with fortitude in the point that as fortitude tends to something arduous and difficult, so also does magnificence: wherefore seemingly it is seated, like fortitude, in the irascible. Yet magnificence falls short of fortitude, in that the \colorbox{red!20}{ardous} thing to which fortitude tends derives its difficulty from a danger that threatens the person, whereas the arduous thing to which magnificence tends, derives its difficulty from the \colorbox{red!20}{disposession} of one's property, which is of much less account than danger to one's person. Wherefore magnificence is accounted a part of fortitude.<lb>
Reply to Objection 1: Justice regards operations in themselves, as viewed under the aspect of something due: but liberality and magnificence regard \colorbox{red!20}{sumptury} operations as related to the passions of the soul, \colorbox{red!20}{albet} in different ways. For liberality regards expenditure in reference to the love and desire of money, which are passions of the \colorbox{red!20}{concupiscibe} faculty, and do not hinder the<lb>
2581<lb>"} \\
        \hline        
\end{longtable}
\endgroup

\end{document}